\documentclass[10pt,journal,compsoc]{IEEEtran}

\usepackage{graphicx}
\usepackage{amsmath}
\usepackage{amssymb}
\usepackage[hyphens]{url}
\usepackage{multirow}
\usepackage{algorithm}
\usepackage{algorithmic}
\usepackage{xcolor}
\usepackage{colortbl}
\usepackage[caption=false]{subfig}
\usepackage{cite}
\usepackage{makecell}
\usepackage{overpic}
\usepackage{rotating,soul}
\usepackage{bbding}
\usepackage{ragged2e}
\usepackage{booktabs}
\usepackage[normalem]{ulem}
\usepackage[breaklinks=true,colorlinks,citecolor=blue]{hyperref}

\usepackage{silence}
\graphicspath{{./Imgs/},{./Imgs/ErrorAnalysis/}}
\DeclareGraphicsExtensions{.pdf,.jpg,.png}

\newcommand{\myPara}[1]{\vspace{.12in}\noindent\textbf{#1}\quad}
\definecolor{mygray}{gray}{.92}

\newcommand{\figref}[1]{Fig.~\ref{#1}}
\newcommand{\tabref}[1]{Table~\ref{#1}}
\newcommand{\equref}[1]{Eq.~\ref{#1}}
\newcommand{\secref}[1]{$\S$\ref{#1}}

\def\ie{\emph{i.e.}}
\def\eg{\emph{e.g.}}

\def\etc{\emph{etc}}
\def\etal{{\em et al.~}}

\begin{document}

\title{Revisiting Computer-Aided Tuberculosis Diagnosis}

\author{Yun Liu, Yu-Huan Wu, Shi-Chen Zhang, Li Liu, Min Wu, and Ming-Ming Cheng
\IEEEcompsocitemizethanks{%
\IEEEcompsocthanksitem This work was mostly done when 
Y. Liu was with VCIP, Nankai University, Tianjin, China. 
He is now with the Institute for Infocomm Research (I2R), 
A*STAR, Singapore. 
(E-mail: vagrantlyun@gmail.com)
\IEEEcompsocthanksitem Y.-H.~Wu is with the Institute of 
High Performance Computing (IHPC), A*STAR, Singapore. 
(E-mail: wu\_yuhuan@ihpc.a-star.edu.sg)
\IEEEcompsocthanksitem S.-C. Zhang and M.-M.~Cheng are with 
VCIP, Nankai University, Tianjin, China. 
(E-mail: zhangshichen@mail.nankai.edu.cn and cmm@nankai.edu.cn)
\IEEEcompsocthanksitem L. Liu is with the College of Electronic Science and Technology,
National University of Defense Technology, Changsha, Hunan, China. 
(E-mail: liuli\_nudt@nudt.edu.cn)
\IEEEcompsocthanksitem M. Wu is with the Institute for Infocomm Research (I2R), A*STAR, Singapore. (E-mail: wumin@i2r.a-star.edu.sg)
\IEEEcompsocthanksitem Corresponding author: Li Liu. (E-mail: liuli\_nudt@nudt.edu.cn)
\IEEEcompsocthanksitem A preliminary version of this work has been published on CVPR (oral) \cite{liu2020rethinking}.
}}

\markboth{IEEE TRANSACTIONS ON PATTERN ANALYSIS AND MACHINE INTELLIGENCE}%
{IEEE TRANSACTIONS ON PATTERN ANALYSIS AND MACHINE INTELLIGENCE}

\IEEEtitleabstractindextext{
\begin{abstract}\justifying
Tuberculosis (TB) is a major global health threat, causing millions of deaths annually.
Although early diagnosis and treatment can greatly improve the chances of survival, it remains a major challenge, especially in developing countries.
Recently, computer-aided tuberculosis diagnosis (CTD) using deep learning has shown promise, but progress is hindered by limited training data.
To address this, we establish a large-scale dataset, namely the Tuberculosis X-ray (TBX11K) dataset, which contains 11,200 chest X-ray (CXR) images with corresponding bounding box annotations for TB areas.
This dataset enables the training of sophisticated detectors for high-quality CTD.
Furthermore, we propose a strong baseline, SymFormer, for simultaneous CXR image classification and TB infection area detection.
SymFormer incorporates Symmetric Search Attention (SymAttention) to tackle the \textit{bilateral symmetry property} of CXR images for learning discriminative features.
Since CXR images may not strictly adhere to the bilateral symmetry property, we also propose Symmetric Positional Encoding (SPE) to facilitate SymAttention through feature recalibration.
To promote future research on CTD, we build a benchmark by introducing evaluation metrics, evaluating baseline models reformed from existing detectors, and running an online challenge.
Experiments show that SymFormer achieves state-of-the-art performance on the TBX11K dataset.
The data, code, and models will be released at \url{https://github.com/yun-liu/Tuberculosis}.
\end{abstract}
}

\maketitle

\IEEEdisplaynontitleabstractindextext
\IEEEpeerreviewmaketitle

\section{Introduction}
\IEEEPARstart{T}{uberculosis} (\textbf{TB}), a pervasive infectious disease, has persistently ranked as the second leading cause of morbidity and mortality, typically following HIV, over the centuries \cite{world2015world, world2017world}. 
Despite the global COVID-19 outbreak in 2020, TB continues to afflict 10 million individuals and accounts for the death of 1.4 million people annually \cite{world2020world}, rendering it the second most lethal infectious disease after COVID-19. 
Principally targeting the respiratory system, TB is caused by Mycobacterium tuberculosis and propagates through sneezing, severe coughing, or other means of disseminating infectious bacteria. 
Hence, TB typically occurs in the lungs through the respiratory tract.
The vulnerability of immunocompromised individuals, including those with HIV and malnourished persons in developing countries, has exacerbated this issue.

The mortality rate among TB patients remains exceedingly high in the absence of appropriate treatment. 
Nevertheless, early diagnosis of TB can significantly increase the recovery rate with the administration of corresponding antibiotics \cite{jaeger2013automatic,chauhan2014role,hwang2016novel}. 
As TB propagates rapidly, early diagnosis also plays a crucial role in controlling the spread of infection \cite{chauhan2014role}. 
The rise of multidrug-resistant TB underscores the urgent need for timely and accurate diagnostic methods to monitor the progress of clinical treatment \cite{gandhi2010multidrug}. 
However, TB diagnosis continues to pose a significant challenge \cite{lopes2017pre,jaeger2013automatic,candemir2013lung,chauhan2014role,hwang2016novel,andersen2000specific,bekmurzayeva2013tuberculosis}.
Specifically, the \textit{gold standard} for TB diagnosis entails the microscopic examination of sputum samples and bacterial cultures to identify Mycobacterium tuberculosis \cite{andersen2000specific,bekmurzayeva2013tuberculosis}. 
To ensure the safety of the examination process, a biosafety level-3 (BSL-3) laboratory is required for culturing Mycobacterium tuberculosis. 
This procedure can typically take several months \cite{andersen2000specific,bekmurzayeva2013tuberculosis,jaeger2013automatic}. Compounding the issue, many hospitals in developing countries and resource-constrained communities \textit{lack the necessary infrastructure} to establish BSL-3 facilities.

On the other hand, X-ray imaging is the most prevalent and data-intensive screening method in current medical image examinations. 
Chest X-ray (\textbf{CXR}) can swiftly detect lung abnormalities caused by pulmonary TB, making it a widely-used tool for TB screening. 
The World Health Organization also recommends CXR as the initial step in TB screening \cite{world2016chest}. 
Early diagnosis through CXR significantly aids in early TB detection, treatment, and prevention of the disease's spread \cite{world2016chest,van2005role,konstantinos2010testing,jaeger2013automatic,candemir2013lung}. 
However, even experienced radiologists may fail to identify TB infections in CXR images, as the human eye struggles to discern TB areas in CXR images due to its limited sensitivity to many details. 
Our human study reveals that experienced radiologists from top hospitals achieve \textit{an accuracy of only 68.7\%} when compared with the gold standard.

Thanks to the remarkable representation learning capabilities, deep learning has outperformed humans in various domains such as face recognition \cite{sun2014deep}, image classification \cite{he2015delving}, object detection \cite{he2017mask,xu2023learning}, edge detection \cite{liu2019richer,liu2022semantic}, and medical image analysis \cite{li2023machine,qiu2023pre,wang2023review}. 
It is reasonable to anticipate the application of deep learning's robust potential to TB diagnosis using CXR. 
Deep learning can automatically localize the precise TB infection site 24 hours a day, never getting tired like humans. 
However, deep learning relies on extensive training data, which cannot be provided by existing TB datasets, as shown in \tabref{tab:dataset}. 
Since it is challenging to collect large-scale TB CXR data due to the high cost and privacy considerations, existing TB datasets have only a few hundred CXR images.
The scarcity of publicly available CXR data has \textit{hindered} the successful application of deep learning in improving computer-aided tuberculosis diagnosis (\textbf{CTD}) performance.

\begin{table}[!t]
  \centering
  \setlength\tabcolsep{4.2pt}
  \caption{\textbf{Summary of publicly available TB datasets.} The size of our dataset is about $17\times$ larger than that of the previous largest dataset. Besides, our dataset annotates TB infection areas with bounding boxes, instead of only image-level labels.}
  \label{tab:dataset}
  \begin{tabular}{c|c|c|c|c} \toprule
    Datasets & Pub. Year & \#Classes & Annotations & \#Samples \\ \midrule
    MC \cite{jaeger2014two} & 2014 & 2 & Image-level & 138 \\
    Shenzhen \cite{jaeger2014two} & 2014 & 2 & Image-level & 662 \\
    DA \cite{chauhan2014role} & 2014 & 2 & Image-level & 156 \\
    DB \cite{chauhan2014role} & 2014 & 2 & Image-level & 150 \\ \midrule
    TBX11K (Ours) & - & 4 & Bounding box & 11,200 \\ \bottomrule
  \end{tabular}
\end{table}

In order to deploy the CTD system to assist TB patients worldwide, it is first necessary to address the issue of insufficient data. 
In this paper, we contribute a large-scale \textbf{Tuberculosis X-ray (TBX11K)} dataset to the community through long-term collaboration with major hospitals. 
This new TBX11K dataset surpasses previous CTD datasets in several aspects: 
i) Unlike previous public datasets \cite{jaeger2014two,chauhan2014role} containing only tens or hundreds of CXR images, TBX11K consists of 11,200 CXR images, approximately 17 times larger than the existing largest dataset, \ie, the Shenzhen dataset \cite{jaeger2014two}, making it feasible to train deep networks; 
ii) In contrast to image-level annotations in previous datasets, TBX11K employs bounding box annotations for TB infection areas, allowing future CTD methods to recognize TB manifestations and detect TB regions for assisting radiologists in definitive diagnoses;
iii) TBX11K comprises four categories: healthy, sick but non-TB, active TB, and latent TB, as opposed to binary classification in previous datasets (\ie, TB or non-TB), enabling future CTD systems to adapt to more complex real-world scenarios and provide people with more detailed disease analyses. 
Each CXR image in the TBX11K dataset is tested using the gold standard (\ie, diagnostic microbiology) of TB diagnosis and annotated by experienced radiologists from major hospitals.
The TBX11K dataset has been de-identified by data providers and exempted by relevant institutions, allowing it to be publicly available to promote future CTD research.

Based on our TBX11K dataset, we propose a simple yet effective framework for CTD, termed as \textbf{SymFormer}.
Inspired by the inherent \textit{bilateral symmetry property} observed in CXR images, SymFormer leverages this property to enhance the interpretation of CXR images.
The bilateral symmetry property denotes the similarity or identical appearance of the left and right sides of the chest, indicating a symmetric pattern. 
This property proves valuable in improving the interpretation of CXR images. 
For instance, if there is a mass or consolidation present on one side of the chest but not the other, it could indicate a problem in that area.
To tackle this property, SymFormer incorporates the novel \textbf{Symmetric Search Attention (SymAttention)} for learning discriminative features from CXR images.
Since CXR images may not strictly be bilaterally symmetric, we also propose the \textbf{Symmetric Positional Encoding (SPE)} to facilitate SymAttention through feature recalibration.
SymFormer conducts simultaneous CXR image classification and TB infection area detection by adding a classification head onto the TB infection area detector with a two-stage training diagram.

To promote future research on CTD, we establish a benchmark on our TBX11K dataset.
Specifically, we adapt the evaluation metrics for image classification and object detection to CTD, which would standardize the evaluation of CTD.
We also launch an online challenge using the test data of TBX11K by keeping the ground truth of the test data private, which would make future comparisons on CTD fair.
Besides, we construct several strong baseline models for CTD by reforming existing popular object detectors.
Extensive comparisons demonstrate the superiority of SymFormer over these baselines.

Compared with the preliminary conference version \cite{liu2020rethinking}, we make plentiful extensions by proposing a novel SymFormer framework for CTD and validating its effectiveness with extensive experiments.
In summary, the contributions of this paper are three-fold:
\begin{itemize}
\item We establish a large-scale CTD dataset, TBX11K, which is much larger, better annotated, and more realistic than previous TB datasets, enabling the training of deep neural networks for simultaneous multi-class CXR image classification and TB infection area detection rather than only binary CXR classification in previous TB datasets.
\item We propose a simple yet effective framework for CTD, namely SymFormer, consisting of the novel Symmetric Search Attention (SymAttention) and Symmetric Positional Encoding (SPE) to leverage the \textit{bilateral symmetry property} of CXR images for significantly improving CTD over baseline models.
\item We build a CTD benchmark on our TBX11K dataset by introducing the evaluation metrics, evaluating several baselines reformed from existing object detectors, and running an online challenge, which is expected to set a good start for future research.
\end{itemize}

\section{Related Work}
In this section, we first revisit previous TB datasets, followed by a review of the existing research on CTD. Since our proposed CTD method SymFormer uses self-attention of vision transformers, we also discuss the recent progress of vision transformers in medical imaging.

\subsection{Tuberculosis Datasets}
Since TB data are very private and it is difficult to diagnose TB with the golden standard, the publicly available TB datasets are very limited.
We provide a summary for the publicly available TB datasets in \tabref{tab:dataset}.
Jaeger \etal \cite{jaeger2014two} established two CXR datasets for TB diagnosis.
The Montgomery County chest X-ray set (MC) \cite{jaeger2014two} was collected through cooperation with the Department of Health and Human Services, Montgomery County, Maryland, USA.
MC dataset consists of 138 CXR images, 80 of which are healthy cases and 58 are cases with manifestations of TB.
The Shenzhen chest X-ray set (Shenzhen) \cite{jaeger2014two} was collected through cooperation with Shenzhen No. 3 People’s Hospital, Guangdong Medical College, Shenzhen, China.
The Shenzhen dataset is composed of 326 norm cases and 336 cases with manifestations of TB, leading to 662 CXR images in total.
Chauhan \etal \cite{chauhan2014role} built two datasets, namely DA and DB, which were obtained from two different X-ray machines at the National Institute of Tuberculosis and Respiratory Diseases, New Delhi.
DA is composed of the training set (52 TB and 52 non-TB CXR images) and the independent test set (26 TB and 26 non-TB CXR images).
DB contains 100 training CXR images (50 TB and 50 non-TB) and 50 test CXR images (25 TB and 25 non-TB).
Note that all these four datasets are annotated with image-level labels for binary CXR image classification.

These datasets are too small to train deep neural networks, so recent research on CTD has been hindered although deep learning has achieved numerous success stories in the computer vision community.
On the other hand, the existing datasets only have image-level annotations, and thus we cannot train TB detectors with previous data.
To help radiologists make accurate diagnoses, we are expected to detect the TB infection areas, not only an image-level classification.
Therefore, the lack of TB data has prevented deep learning from bringing success to practical CTD systems that have the potential to save millions of TB patients every year.
In this paper, we build a large-scale dataset with bounding box annotations for training deep neural networks for simultaneous CXR image classification and TB infection area detection.
The presentation of this new dataset is expected to benefit future research on CTD and promote more practical CTD systems.

\subsection{Computer-aided Tuberculosis Diagnosis}
Owing to the lack of data, traditional CTD methods cannot train deep neural networks.
Thus, traditional methods mainly use hand-crafted features and train binary classifiers for CXR image classification.
Jaeger \etal \cite{jaeger2013automatic} first segmented the lung region using a graph cut segmentation method \cite{boykov2006graph}.
Then, they extracted hand-crafted texture and shape features from the lung region.
Finally, they apply a binary classifier, \ie, support vector machine (SVM), to classify the CXR image as normal or abnormal.
Candemir \etal \cite{candemir2013lung} adopted image retrieval-based patient-specific adaptive lung models to a nonrigid registration-driven robust lung segmentation method, which would be helpful for traditional lung feature extraction \cite{jaeger2013automatic}.
Chauhan \etal \cite{chauhan2014role} implemented a MATLAB toolbox, TB-Xpredict, which adopted Gist \cite{oliva2006building} and PHOG \cite{bosch2007representing} features for the discrimination between TB and non-TB CXR images without requiring segmentation \cite{cheng2016hfs,liu2018deep}.
Karargyris \etal \cite{karargyris2016combination} extracted shape features to describe the overall geometrical characteristics of lungs and texture features to represent image characteristics.

Instead of using hand-crafted features, Lopes \etal \cite{lopes2017pre} adopted the frozen convolutional neural networks pre-trained on ImageNet \cite{deng2009imagenet} as the feature extractors for CXR images.
Then, they train SVM to classify the extracted deep features.
Hwang \etal \cite{hwang2016novel} trained an AlexNet \cite{krizhevsky2012imagenet} for binary classification (TB and non-TB) using a private dataset.
Other private datasets are also used in \cite{lakhani2017deep} for image classification networks.
However, our proposed large-scale dataset, \ie, TBX11K, has been made publicly available to promote research in this field.
With our new dataset, we propose a transformer-based CTD method, SymFormer, for simultaneous CXR image classification and TB infection area detection, which serves as a strong baseline for future research on CTD by achieving state-of-the-art performance.

\subsection{Vision Transformers in Medical Imaging}
Transformer \cite{vaswani2017attention} is initially introduced in natural language processing (NLP), and it has a good ability to capture long-range dependencies.
Pioneering works on adapting transformers to vision tasks, such as ViT \cite{dosovitskiy2021image}, DeiT \cite{touvron2021training}, and P2T \cite{wu2022p2t}, showed that transformer networks can surpass the widely-used convolutional neural networks. 
Therefore, vision transformers attract increasing attention from the computer vision community, including medical imaging.
Various efforts have been made to incorporate vision transformers into medical image segmentation \cite{zhang2021transfuse,xie2021cotr,ji2021multi,gao2021utnet,tao2022spine} and medical image classification \cite{shao2021transmil,park2021federated,mo2023hover,bhattacharya2022radiotransformer,park2022multi}.
However, the adoption of transformer-based techniques for medical image detection lags behind that of segmentation and classification.

Most studies utilizing vision transformers for medical image detection are primarily built on the detection transformer (DETR) framework \cite{carion2020end}.
The pioneering work in this field is COTR \cite{shen2021cotr}, comprising a convolutional neural network for feature extraction, hybrid convolution-in-transformer layers for feature encoding, transformer decoder layers for object querying, and a feed-forward network for polyp detection.
Mathai \etal \cite{mathai2022lymph} employed DETR \cite{carion2020end} to detect lymph nodes in T2 MRI scans, which can be used to evaluate lymphoproliferative diseases.
Li \etal \cite{li2022satr} proposed a Slice Attention Transformer (SATr) block to model the long-range dependency among different computed tomography (CT) slices, which can be plugged into convolution-based models for universal lesion detection.
Please refer to recent survey papers \cite{shamshad2022transformers,he2023transformers,li2023transforming} for a more comprehensive review of vision transformers in medical imaging.
In this paper, we propose SymFormer for CTD using CXR images.
SymFormer conducts simultaneous CXR image classification and TB infection area detection.
It leverages SymAttention to tackle the \textit{bilateral symmetry property} of CXR images, which is further promoted by SPE.
With SymAttention and SPE, SymFormer exhibits much better performance than recent popular object detector baselines, suggesting its superiority in CTD.

\begin{figure}[!t]
\centering
\includegraphics[width=.421\linewidth]{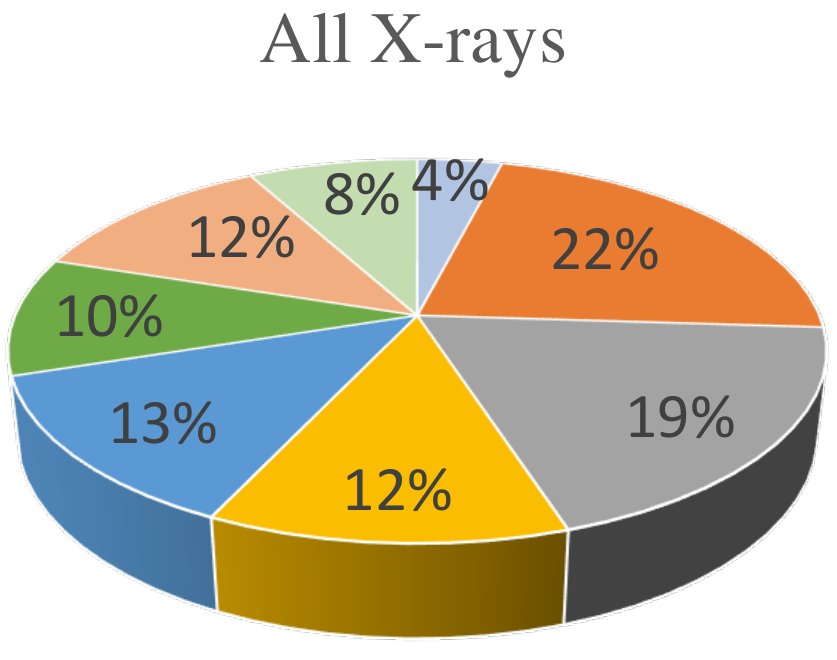} \hfill
\includegraphics[width=.565\linewidth]{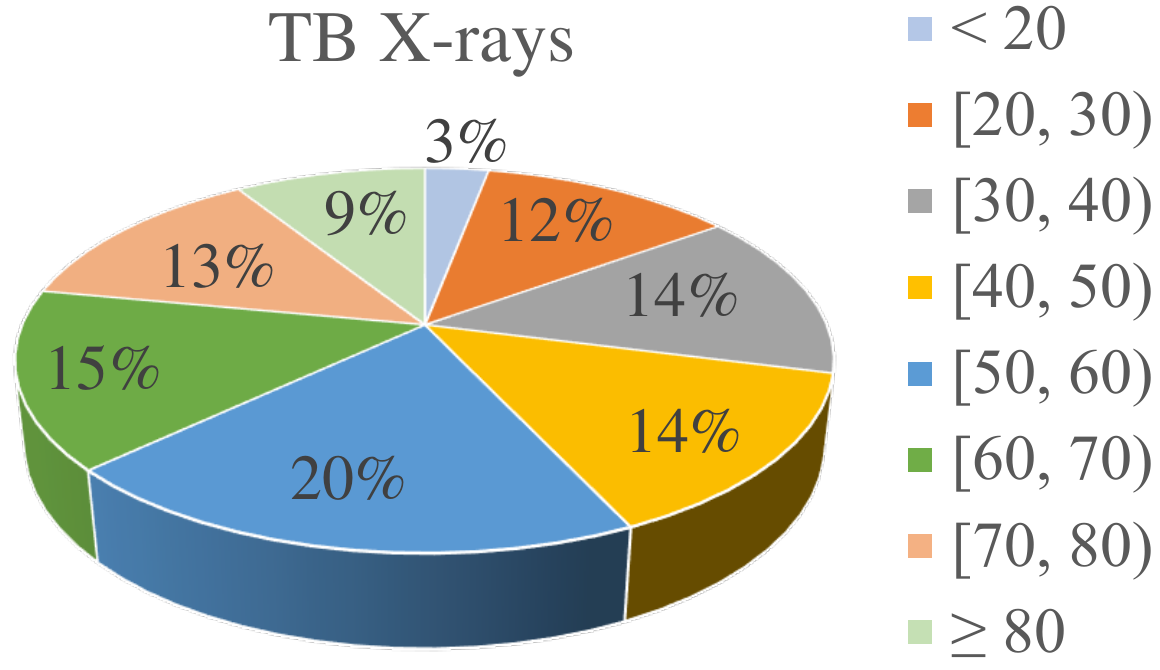}
\caption{Age distributions for the entire TBX11K dataset and specifically for TBX11K TB X-rays.}
\label{fig:age}
\end{figure}

\begin{figure}[!t]
\centering
\includegraphics[width=.44\linewidth]{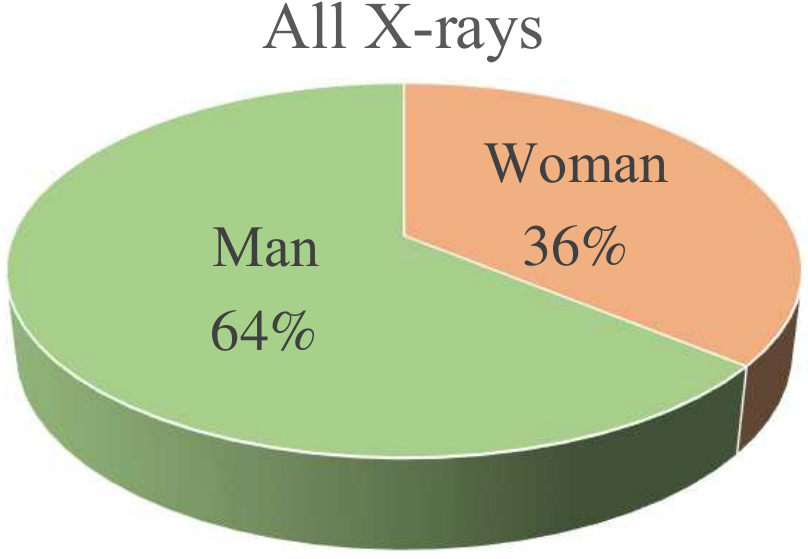} \hfill
\includegraphics[width=.45\linewidth]{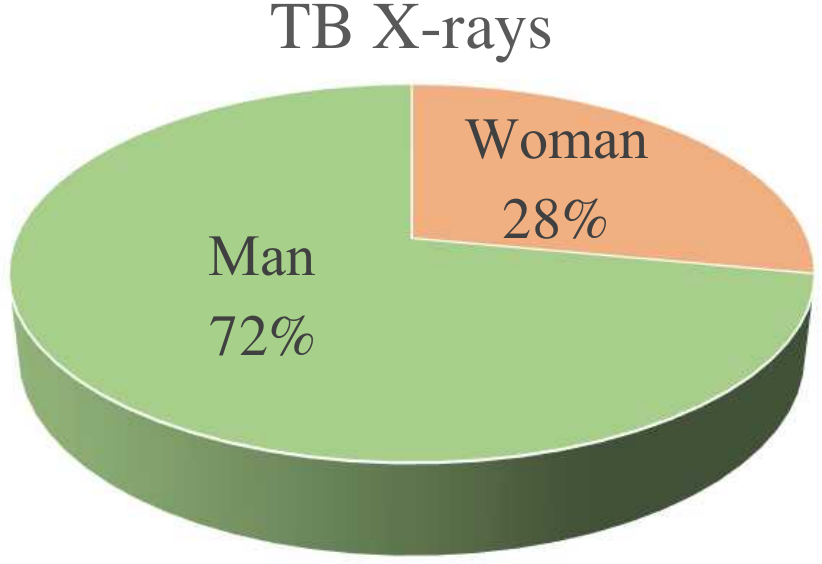}
\caption{Gender distributions for the entire TBX11K dataset and specifically for TBX11K TB X-rays.}
\label{fig:gender}
\end{figure}

\section{TBX11K Dataset}
Deep neural networks are highly dependent on large amounts of training data, while existing public TB datasets are not large-scale as shown in \tabref{tab:dataset}.
To address this issue, we establish a comprehensive and large-scale dataset called TBX11K, which enables the training of deep networks for CTD.
In this section, we first describe how we collect and annotate the CXR data in \secref{sec:data}. 
Next, we present the results of a human study conducted by experienced radiologists in \secref{sec:human}.
Finally, we discuss potential research topics that can be explored using our TBX11K dataset in \secref{sec:topics}.

\subsection{Data Collection and Annotation} \label{sec:data}
To collect and annotate the data, we adhere to four primary steps: i) establishing a taxonomy, ii) collecting CXR data, iii) professional data annotation, and iv) dataset splitting. We will introduce each of these steps in detail below.

\subsubsection{Taxonomy Establishment}
The current TB datasets only consist of two categories: TB and non-TB, where non-TB refers to healthy cases. However, in practice, abnormalities in CXR images that indicate TB, atelectasis, cardiomegaly, effusion, infiltration, mass, nodule, \etc., have similar abnormal patterns such as blurry and irregular lesions, which differ significantly from healthy CXR that have almost clear patterns. Therefore, relying solely on healthy CXR as the negative category leads to biases that can cause large false positives in the model's prediction for clinical scenarios where there are many sick but non-TB patients. To address this issue and promote the practical application of CTD, we propose a new category, sick but non-TB, in our dataset.
Furthermore, differentiating between active TB and latent TB is crucial in providing patients with proper treatment. Active TB results from Mycobacterium TB infection or reactivation of latent TB, while individuals with latent TB are neither sick nor contagious. Therefore, we have divided TB into two categories of active TB and latent TB in our dataset.
In light of the above analysis, the proposed TBX11K dataset includes four categories: healthy, sick but non-TB, active TB, and latent TB.

\begin{figure}[!t]
\centering
\includegraphics[width=\linewidth]{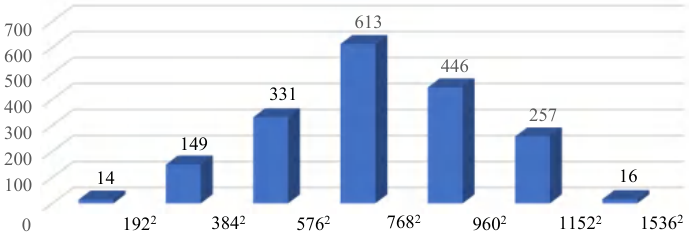}
\caption{\textbf{Distribution of the sizes of TB bounding boxes in the TBX11K dataset.} Each bin represents a specific range of bounding box areas. The left and right values of each bin correspond to its area range, and the height of the bin represents the number of TB bounding boxes within that range. It should be noted that the CXR images in TBX11K have a resolution of about $3000 \times 3000$.}
\label{fig:box_areas}
\end{figure}

\subsubsection{Data Collection}
The collection of TB CXR data presents two main challenges: i) The high privacy of CXR data, particularly TB CXR data, making it almost impossible for individuals to access the raw data without risking breaking the law; ii) The scarcity of definitively tested TB CXR images, due to the complex and lengthy process of examining Mycobacterium TB using the golden standard \cite{andersen2000specific,bekmurzayeva2013tuberculosis}, despite the millions of TB patients worldwide.
To address these challenges, we collaborate with top hospitals in China to gather the CXR data. Our resulting TBX11K dataset comprises 11,200 CXR images, including 5,000 healthy cases, 5,000 sick but non-TB cases, and 1,200 TB cases. Each CXR image corresponds to a unique individual.
Of the 1,200 TB CXR images, 924 are active TB cases, 212 are latent TB cases, 54 contain both active and latent TB, and 10 are uncertain cases whose TB types cannot currently be recognized. We include 5,000 sick but non-TB cases to cover a broad range of radiograph diseases that can appear in clinical scenarios.
The data providers have de-identified the data, and relevant government institutions have exempted the dataset, making it publicly available legally.

All CXR images are in a resolution of approximately $3000 \times 3000$. Each CXR image is accompanied by corresponding age and gender information, providing comprehensive clinical clues for the diagnosis of TB.
The age distribution and gender distribution are presented in \figref{fig:age} and \figref{fig:gender}, respectively. 
From \figref{fig:age}, it is evident that 60\% of the TB patients fall within the age range of 20 to 60 years. Only a small percentage of young individuals (aged less than 20) are infected with TB, specifically 3\% of TB patients. This finding is consistent with recent medical research \cite{byng2016does,garcia2018adolescents,dong2022age}.
\figref{fig:gender} illustrates that the majority of TB patients are male, aligning with clinical observations that TB is more prevalent in men than women globally \cite{nhamoyebonde2014biological,horton2016sex}.
We have annotated TB infection areas using bounding boxes (introduced in \secref{sec:anno}), and the distribution of the sizes of these TB bounding boxes is depicted in \figref{fig:box_areas}. As evident from the figure, TB infection areas exhibit a wide range of sizes, manifesting varying degrees of TB severity.
Taken together with the previous analyses on taxonomy, category distribution, age distribution, gender distribution, and TB infection size distribution, we can come to the conclusion that the new TBX11K dataset is \textit{both representative of the general population and consistent with real-world clinical scenarios.}

\subsubsection{Professional Data Annotation} \label{sec:anno}
Our dataset comprises CXR images that have undergone rigorous testing using the golden standard, which provides image-level labels. 
However, while this approach enables us to categorize a CXR image as indicative of TB if the sputum of the corresponding patient shows manifestations of the disease, it does not reveal the specific location or extent of the TB in the CXR image. 
The ability to detect these TB infection areas is crucial to enable radiologists to make informed decisions.
Currently, relying solely on image-level predictions makes it difficult for the human eye to identify TB infection areas, as evidenced by the low accuracy of radiologists during clinical examinations (see \secref{sec:human}). 
By simultaneously providing image classification and TB localization results, CTD systems have the potential to enhance the accuracy and efficiency of radiologists in making informed decisions.

To achieve our goal, our TBX11K dataset includes bounding box annotations for TB infection areas in CXR images.
To the best of our knowledge, this is the first dataset designed for TB infection area detection.
These annotations are carried out by experienced radiologists from top hospitals.
Specifically, each TB CXR image in the dataset is first labeled by a radiologist with 5-10 years of experience in TB diagnosis. 
Subsequently, another radiologist with over 10 years of experience in TB diagnosis reviews the box annotations. 
The radiologists do not just label bounding boxes for TB areas but also identify the type of TB (active or latent) for each box. 
To ensure consistency, the labeled TB types are double-checked against the image-level labels produced by the golden standard.
In the event of a mismatch, the CXR image is placed in the unlabeled data for re-annotation, and the annotators do not know which CXR image was previously labeled incorrectly. 
If a CXR image is labeled incorrectly twice, we inform the annotators of the gold standard for that CXR image and request that they discuss how to re-annotate it.
This double-checked process ensures that the annotated bounding boxes are highly reliable for TB infection area detection. 
Additionally, non-TB CXR images are only labeled with image-level labels produced by the golden standard. 
Examples of the TBX11K dataset are shown in \figref{fig:vis}, and the distribution of TB bounding box sizes is displayed in \figref{fig:box_areas}, indicating that most TB bounding boxes are in the range of $(384^2, 960^2]$.

\begin{table}[!t]
  \centering
  \setlength\tabcolsep{5.5pt}
  \caption{\textbf{Split for the TBX11K dataset.} ``Active \& Latent TB'' refers to CXR images with both active and latent TB; ``Active TB'' refers to CXR images with only active TB; ``Latent TB'' refers to CXR images with only latent TB; ``Uncertain TB'' refers to TB CXR images where the type of TB infection cannot be recognized using current medical conditions.}
  \label{tab:split}
  \begin{tabular}{c|c|c|c|c|c} \toprule
    & Classes & Train & Val & Test & Total \\ \midrule
    \multirow{2}{*}{Non-TB} & Healthy & 3,000 & 800 & 1,200 & 5,000 \\
    & Sick \& Non-TB & 3,000 & 800 & 1,200 & 5,000 \\ \midrule
    \multirow{4}{*}{TB} & Active TB &  473 & 157 & 294 & 924 \\
    & Latent TB & 104 & 36 & 72 & 212 \\ 
    & Active \& Latent TB & 23 & 7 & 24 & 54 \\
    & Uncertain TB & 0 & 0 & 10 & 10 \\ \midrule
    \multicolumn{2}{c|}{Total} & 6,600 & 1,800 & 2,800 & 11,200 
    \\ \bottomrule
  \end{tabular}%
\end{table}

\subsubsection{Dataset Splitting}
We have partitioned the data into three subsets: training, validation, and test, following the split detailed in \tabref{tab:split}. 
The ground truths for both the training and validation sets have been made public, whereas the ground truths for the test set remain confidential. This is because we have launched an online challenge using the test data on our \href{https://codalab.lisn.upsaclay.fr/competitions/7916}{website}.
To ensure a more representative dataset, we have considered four distinct TB cases: i) CXR images with active TB only, ii) CXR images with latent TB only, iii) CXR images with both active and latent TB, and iv) CXR images with uncertain TB type that cannot be recognized under current medical conditions. 
For each TB case, we have maintained a ratio of $3:1:2$ for the number of TB CXR images in the training, validation, and test sets. 
It is worth noting that the uncertain TB CXR images have been assigned to the test set, enabling researchers to evaluate class-agnostic TB detection using these 10 uncertain CXR images. 
We recommend that researchers train their models on the training set, tune hyper-parameters on the validation set, and report the model's performance on the test set after retraining using the union of the training and validation sets. 
This approach follows scientific experiment settings and is expected to yield reliable results.

\subsection{Human Study by Radiologists} \label{sec:human}
The human study involving radiologists is a critical component in understanding the role of CTD in clinical settings. 
We begin by randomly selecting 400 CXR images from the test set of the new TBX11K dataset, which includes 140 healthy CXR images, 140 sick but non-TB CXR images, and 120 CXR images with TB. 
Of the 120 CXR images with TB, 63 show active TB, 41 show latent TB, 15 show both active and latent TB, and 1 shows uncertain TB.
Next, we invite an experienced radiologist from a major hospital with over 10 years of work experience to label the CXR images according to four image-level categories: healthy, sick but non-TB, active TB, and latent TB. 
If a CXR image displays both active and latent TB manifestations, the radiologist assigns both labels. 
It is important to note that this radiologist is different from those who labeled the original dataset.

The radiologist achieves an accuracy of only 68.7\% when compared to the ground truth produced by the golden standard. 
If we ignore the differentiation between active and latent TB, the accuracy improves to 84.8\%, but distinguishing between the types of TB is crucial for effective clinical treatment. 
This low performance highlights one of the major challenges in TB diagnosis, treatment, and prevention. 
Unlike natural color images, CXR images are grayscale and often have fuzzy and blurry patterns, making accurate recognition challenging.
Unfortunately, diagnosing TB with the golden standard can take several months in a BSL-3 laboratory \cite{andersen2000specific,bekmurzayeva2013tuberculosis}, which is not feasible in many parts of the world.
The challenge in TB diagnosis leads to TB becoming the second most common infectious disease worldwide after HIV.
However, we will show in our upcoming study that deep-learning-based CTD models trained on the proposed TBX11K dataset can significantly outperform even experienced radiologists, offering hope for improved TB diagnosis and treatment.

\subsection{Potential Research Topics} \label{sec:topics}
Moving forward, we discuss some potential research topics related to CTD using our newly developed TBX11K dataset.

\myPara{Simultaneous classification and detection.}
Our TBX11K dataset opens up new possibilities for conducting research on CTD, including CXR image classification and TB infection area detection. 
Our test set includes a broad range of health and non-TB sick data, enabling the simulation of clinical data distribution for evaluating CTD systems.
We believe that the development of simultaneous CXR image classification and TB infection area detection systems would be a challenging and fascinating research topic, with potential applications for assisting radiologists in TB diagnosis. 
Deploying such systems could ultimately improve the accuracy and efficiency of TB diagnosis and treatment.

\myPara{Imbalanced data distribution.}
In addition to the challenge of simultaneous detection and image classification, our TBX11K dataset also presents an imbalanced data distribution across different categories. 
However, we believe that this data imbalance is reflective of real-world clinical scenarios.
When patients undergo chest examinations, they may be experiencing discomfort or illness, increasing the likelihood of getting sick, and our dataset captures this reality with only 44.6\% of takers being healthy.
TB is just one of many possible chest diseases, and our dataset reflects this reality with only 10.7\% of takers being infected with TB, while 44.6\% are sick but non-TB.
Latent TB can result from two scenarios: exposure to active TB and conversion from active TB after treatment. 
Most cases of latent TB are caused by exposure to active TB. 
However, individuals with latent TB are not sick or contagious and are unlikely to seek medical attention, resulting in a higher number of active TB cases in our dataset than latent TB cases.
This data imbalance presents a challenge for future CTD methods, which must be designed to overcome this problem in practice. 
For example, methods for training models on the imbalanced TBX11K training set will need to be developed to improve the accuracy of TB diagnosis.

\myPara{Incremental learning with private data.}
Incremental learning is a machine learning technique that involves updating a model's parameters with new data as it becomes available, without requiring the model to be retrained from scratch. 
Given the high privacy concerns surrounding TB CXR data, researchers may possess private data that cannot be released. 
In such cases, it may be beneficial to use a model pre-trained on the TBX11K dataset as the base model. 
Researchers can then leverage incremental learning to fine-tune the pre-trained model using their private data, thereby enhancing the model's capacity for accurate CTD.
Hence, investigating the potential of incremental learning for CTD using the newly developed TBX11K dataset would also be a crucial research direction.

\begin{figure}[!t]
  \centering
  \includegraphics[width=\linewidth]{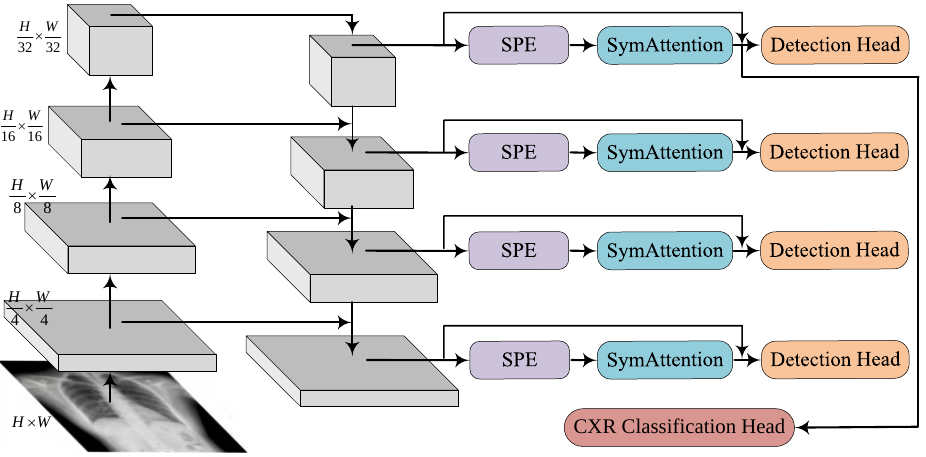}
  \caption{\textbf{Illustration of the proposed SymFormer framework.} FPN \cite{lin2017feature} is applied to generate the feature pyramid.}
  \label{fig:framework}
\end{figure}

\section{Our SymFormer Framework}
In this section, we first present an overview of our SymFormer framework in \secref{sec:overview}. 
Then, we describe our \textbf{Symmetric Abnormity Search (SAS)} method in \secref{sec:sas}. 
SAS consists of two components: \textbf{Symmetric Positional Encoding (SPE)} in \secref{sec:spe}, and \textbf{Symmetric Search Attention (SymAttention)} in \secref{sec:attention}. 
Next, we introduce the TB diagnosis heads for SymFormer in \secref{sec:heads}. 
Finally, we present the two-stage training diagram for simultaneous CXR image classification and TB infection area detection in \secref{sec:training}.

\subsection{Overview} \label{sec:overview}
We illustrate the overall pipeline of SymFormer in \figref{fig:framework}.
SymFormer comprises three parts: feature extraction, symmetric abnormity search, and TB diagnosis heads. 
We will elaborate on each part below.

\myPara{Feature extraction.}
For the sake of convenience, we take ResNets \cite{he2016deep} as an example backbone network for feature extraction due to its generality acknowledged by the community.
When given a CXR image as input, the backbone network outputs features in four stages, which are scaled down by factors of $1/4$, $1/8$, $1/16$, and $1/32$, respectively, in comparison to the input size. 
As the sizes and shapes of TB infection areas vary widely, it is crucial to capture multi-scale features from the backbone network. 
In order to achieve this, a feature pyramid network (FPN) \cite{lin2017feature} is applied upon the backbone network, which generates a feature pyramid, \ie, feature maps at different scales. 
We denote the feature pyramid as $\mathbf{F}=\{\mathbf{F}_1, \mathbf{F}_2, \mathbf{F}_3, \mathbf{F}_4\}$ \textit{w.r.t.} $\mathbf{F}_i \in \mathbb{R}^{C \times \frac{H}{2^{i+1}} \times \frac{W}{2^{i+1}}} (i \in \{1,2,3,4\})$, in which $C$ is the feature dimension and $H$ and $W$ are the height and width of the input CXR image, respectively.
The feature pyramid is effective at enabling TB infection detection across different feature levels.

\myPara{Symmetric abnormity search.}
The SAS module serves to enhance the extracted feature pyramid $\mathbf{F}$. 
To achieve this, an SAS module is incorporated after each side output of FPN \cite{lin2017feature} to process each feature map $\mathbf{F}_i$ in the feature pyramid $\mathbf{F}$.
The enhanced feature pyramid is expressed as $\hat{\mathbf{F}}=\{\hat{\mathbf{F}}_1, \hat{\mathbf{F}}_2, \hat{\mathbf{F}}_3, \hat{\mathbf{F}}_4\}$ \textit{w.r.t.} $\hat{\mathbf{F}}_i \in \mathbb{R}^{C \times \frac{H}{2^{i+1}} \times \frac{W}{2^{i+1}}} (i \in \{1,2,3,4\})$.
The SAS modules at various side outputs share the same weights to reduce the number of network parameters.
According to the \textit{bilateral symmetry property}, the bilaterally symmetric regions in a normal CXR image should look similar or identical.
The SAS module leverages this insight by searching for symmetric positions in each position of the feature map to determine if it is normal. 
The SAS module consists of three components: SPE, SymAttention, and a feed-forward network.
While the CXR image may not be strictly symmetric, the SPE is designed to recalibrate the features, which then benefits the SymAttention for symmetric-search-based feature enhancement.

\myPara{TB diagnosis heads.}
We connect two types of TB diagnosis heads to the feature pyramid $\hat{\mathbf{F}}$, which is enhanced by the SAS module, for performing TB infection area detection and CXR image classification, respectively. 
Each feature map in the feature pyramid $\hat{\mathbf{F}}$ is fed into the detection head, and each detected bounding box is expected to cover a TB infection area.
However, there is a risk of introducing false positives for non-TB CXR images during TB infection area detection, which leads to unnecessary costs for radiologists to check these false positives for clinical diagnosis. 
To address this issue, we feed the feature map $\hat{\mathbf{F}}_4$ at the top level of the enhanced feature pyramid into a classification head to determine whether a CXR image contains TB or not. 
If a CXR image is classified as TB, radiologists can further examine the detected TB infection areas for a more accurate and detailed clinical diagnosis. If a CXR image is classified as non-TB, the detected areas need not be checked further by radiologists.

\subsection{Symmetric Abnormity Search} \label{sec:sas}
Bilateral symmetry is a property of CXR images where the structures on the left and right sides of the chest appear similar or identical. 
In other words, if a line is drawn down the center of the CXR image, the structures on either side of the line should be approximately the same size and shape. 
This property plays a crucial role in the interpretation of CXR images since it enables radiologists and clinicians to identify asymmetries or abnormalities in the lung fields. 
For example, the presence of a mass or consolidation on one side of the lung but not the other could indicate a problem in that area.
However, it is worth noting that perfect bilateral symmetry is not always present in normal CXR images, depending on the patient's pose and position relative to the X-ray machine when the CXR image is taken.

Our proposed method, SAS, leverages the bilateral symmetry property to enhance the feature representations of CXR images. 
As mentioned above, the lungs in CXR images may not be strictly symmetric. 
To account for this, SAS first incorporates SPE for feature recalibration. 
This recalibrated feature map is then used by SymAttention to search the symmetric adjacent area of each spatial position in the feature map, where the symmetric adjacent area refers to the adjacent area of the bilaterally symmetric position for a given position.
SymAttention aggregates features in the symmetric adjacent area in an adaptive way through attention. 
The adjacent area is also determined in a learning way. 
By forcing each spatial position to look at the symmetric adjacent area, as suggested by the bilateral symmetry property, we can learn discriminative features for the CXR image for CTD.

\subsubsection{Symmetric Positional Encoding} \label{sec:spe}
To incorporate positional information into self-attention computations for a feature map, we must add positional encoding to the feature map. 
There are two types of positional encoding: absolute positional encoding and relative positional encoding \cite{vaswani2017attention,dosovitskiy2021image}. 
Our method, called SPE, is based on absolute positional encoding, and our experiments indicate that relative positional encoding is inferior to our SPE, as shown in \secref{sec:ablation}.
The widely-used absolute positional encoding \cite{vaswani2017attention,dosovitskiy2021image} employs sine and cosine functions of different frequencies:
\begin{equation} \label{eq:pos_enc}
\begin{aligned}
    \mathbf{P}[pos, 2j] \hspace{6.1mm} &= \sin (pos / 10000^{\frac{2j}{C}}), \\
    \mathbf{P}[pos, 2j+1] &= \cos (pos / 10000^{\frac{2j}{C}}),
\end{aligned}
\end{equation}
where $pos$ denotes the spatial position and $j$ indexes the feature dimension.
For each input feature map $\mathbf{F}_i$ from the feature pyramid $\mathbf{F}$, we use \equref{eq:pos_enc} to calculate the corresponding positional encoding $\mathbf{P}_i$.
$\mathbf{P}_i$ has the same shape as $\mathbf{F}_i$ so that $\mathbf{P}_i$ and $\mathbf{F}_i$ can be summed.

As mentioned earlier, CXR images may not strictly adhere to the bilateral symmetry property as they can have slight rotations and translations.
The proposed SPE is designed to tackle this issue by feature recalibration.
SPE first splits the positional encoding $\mathbf{P}_i$ into two sides, \ie, $\mathbf{P}_i^{\textit{left}}$ and $\mathbf{P}_i^{\textit{right}}$, by drawing a line down the center of $\mathbf{P}_i$.
Then, we transfer $\mathbf{P}_i^{\textit{right}}$ to the left side using spatial transformer networks (STN) \cite{jaderberg2015spatial} and horizontal flipping.
Finally, we concatenate the transformed left-side positional encoding and $\mathbf{P}_i^{\textit{right}}$ along the $x$ dimension to form the output $\mathbf{P}_i^{\textit{sym}}$.
This process can be formulated as follows:
\begin{equation} \label{eq:spe}
\begin{aligned}
    \mathbf{T}_i \hspace{4mm} &= {\rm STN}(\mathbf{F}_i;\Theta) \\
    \mathbf{P}_i^{\textit{trans}} &= {\rm Flip}_x( \mathcal{T}_\Theta( \mathbf{P}_i^{\textit{right}}; \mathbf{T}_i)), \\
    \mathbf{P}_i^{\textit{sym}} \hspace{1mm} &= {\rm Concat}_x(\mathbf{P}_i^{\textit{trans}}, \mathbf{P}_i^{\textit{right}}),
\end{aligned}
\end{equation}
in which $\Theta$ is the weights of STN; $\mathbf{T}_i$ is the affine transformation matrix; $\mathcal{T}_\Theta$ indicates the affine transformation; ${\rm Flip}_x$ represents horizontal flipping; and ${\rm Concat}_x$ stands for concatenation along the $x$ dimension.
In \equref{eq:spe}, $\mathbf{P}_i^{\textit{right}}$ can be replaced with $\mathbf{P}_i^{\textit{left}}$ by swapping the order of the inputs of ${\rm Concat}_x$.
However, our experiments in \secref{sec:ablation} show that $\mathbf{P}_i^{\textit{right}}$ performs slightly better than $\mathbf{P}_i^{\textit{left}}$.
For each input $\mathbf{F}_i$ ($i \in \{1,2,3,4\}$), we compute the corresponding $\mathbf{P}^{\textit{sym}}_i$ using \equref{eq:spe}.
Using the SPE $\mathbf{P}^{\textit{sym}}_i$, we recalibrate the input feature map through 
\begin{equation}
    \mathbf{F}_i^{\textit{recalib}} = \mathbf{F}_i + \mathbf{P}^{\textit{sym}}_i.
\end{equation}
The output $\mathbf{F}_i^{\textit{recalib}}$ will facilitate the calculation of the subsequent SymAttention.

\myPara{Micro designs of STN.}
The spatial transformation in \equref{eq:spe} is conditional on both the input feature and positional encoding.
We feed the input feature $\mathbf{F}_i$ into a small network of STN to predict the affine transformation matrix $\mathbf{T}_i$, which is then employed for the affine transformation of the one-side positional encoding $\mathbf{P}_i^{\textit{right}}$.
The small network includes two alternating max-pooling and Conv-ReLU layers.
Then, a flattening operation is carried out on the spatial dimension, followed by a multilayer perceptron (MLP) to predict the affine matrix.
We initialize the MLP to ensure that the affine transformation with the initial affine matrix is equivalent to an identical mapping.

\subsubsection{Symmetric Search Attention} 
\label{sec:attention}
Self-attention has gained popularity in various fields \cite{tang2023predicting,wei2023cat,sun2023rethinking,qiu2023boosting} due to its ability to learn relationships among elements within a sequence or image \cite{vaswani2017attention,dosovitskiy2021image}. 
In medical image analysis, self-attention has been applied to identify relevant features in images and enhance disease detection. 
However, classical self-attention performs global relationship modeling by calculating attention weights for each reference location, which fuses features from all locations.
This approach may not be optimal for CTD with CXR images.
Specifically, natural images can be captured in various scenarios and contain various objects and elements, so global relationship modeling is beneficial for the understanding of the entire scene.
However, CXR images only depict the human chest in a single scenario, and the difference among various CXR images is often limited to the presence of elusive abnormity regions.
Therefore, global relationship modeling may be \textit{redundant} for CXR images, limiting the capacity of self-attention to learn relevant relationships for enhancing feature representation.
This is because it is challenging for a neural network to automatically identify a few relevant locations out of thousands of redundant locations.
For instance, in our experiments, we observe that the DETR detection framework \cite{carion2020end} can not converge when used to discriminate indistinguishable TB features in CTD.

To tackle this challenge, we propose SymAttention, which leverages the bilateral symmetry property to aid self-attention in identifying relevant locations in CXR images.
As previously mentioned, radiologists can diagnose TB by comparing the bilaterally symmetric locations of the two sides of the lungs.
Consequently, the relevant locations for each reference location in CXR images are the bilaterally symmetric locations.
Inspired by this, SymAttention searches for features in a symmetrical pattern across the left and right lungs, allowing each reference location only attends to the locations around the bilaterally symmetric location of the reference location.
Given the feature map $\mathbf{F}_i^{\textit{recalib}}$, we first select a small set of key sampling locations, following Deformable DETR \cite{zhu2021deformable}.
Let $K$ denote the number of selected locations, and $M$ denote the number of heads in the self-attention calculation.
The coordinate shifts of the selected locations can be learned by
\begin{equation} \label{eq:coord}
\Delta \mathbf{p}^x_i = \mathbf{W}^{\textit{pos}}_x \mathbf{F}_i^{\textit{recalib}}, \qquad
\Delta \mathbf{p}^y_i = \mathbf{W}^{\textit{pos}}_y \mathbf{F}_i^{\textit{recalib}},
\end{equation}
in which $\mathbf{W}^{\textit{pos}}_x, \mathbf{W}^{\textit{pos}}_y \in \mathbb{R}^{(M\times K)\times C}$ are trainable parameter matrices.
The attention $\mathbf{A}_i$ and value $\mathbf{F}_i^{\textit{v}}$ are simply calculated using
\begin{equation}
\begin{aligned}
\mathbf{A}_i &= {\rm Softmax}({\rm Reshape}(\mathbf{W}^{\textit{att}} \mathbf{F}_i^{\textit{recalib}})), \\
\mathbf{F}_i^{\textit{v}} &= \mathbf{W}^{\textit{value}} \mathbf{F}_i^{\textit{recalib}}
\end{aligned}
\end{equation}
where $\mathbf{W}^{\textit{att}} \in \mathbb{R}^{(M\times K)\times C}$, $\mathbf{W}^{\textit{value}} \in \mathbb{R}^{C\times C}$ are trainable parameter matrices and the softmax function is performed along the dimension of $K$.
Then, we reshape $\mathbf{F}_i^{\textit{v}}$ like
\begin{equation}
\mathbf{F}_i^{\textit{v}} \in \mathbb{R}^{C \times \frac{H}{2^{i+1}} \times \frac{W}{2^{i+1}}} \to \mathbf{F}_i^{\textit{v}} \in \mathbb{R}^{M \times \frac{C}{M} \times \frac{H}{2^{i+1}} \times \frac{W}{2^{i+1}}}.
\end{equation}
Next, SymAttention can be formulated as 
\begin{equation}
\begin{aligned}
\mathbf{F}_i^{\textit{att}} = {\rm Concat}_{m=1}^M ( \sum_{k=1}^K (\mathbf{A}_i[m, k]\cdot \mathbf{F}_i^{\textit{v}}[m, :,
\mathbf{p}^y_i + \Delta \mathbf{p}^y_i[m,k], \\ \uwave{\frac{W}{2^{i+1}} - (\mathbf{p}^x_i + \Delta \mathbf{p}^x_i[m,k]) + 1}])),
\end{aligned}
\end{equation}
in which ${\rm Concat}_{m=1}^M$ means to concatenate all the results generated by setting $m$ from $1$ to $M$. The term with the \uwave{wavy underline} projects the sampled locations onto the bilaterally symmetric locations by taking the vertical centerline as the line of symmetry, which is the core of the proposed SymAttention.
Finally, to ease optimization, a residual connection is connected, followed by an MLP:
\begin{equation} \label{eq:residual}
\begin{aligned}
\hat{\mathbf{F}}_i^{\textit{att}} = \mathbf{W}^{\textit{proj}} \mathbf{F}_i^{\textit{att}} + \mathbf{F}_i, \qquad 
\hat{\mathbf{F}}_i = {\rm MLP}(\hat{\mathbf{F}}_i^{\textit{att}}) + \hat{\mathbf{F}}_i^{\textit{att}},
\end{aligned}
\end{equation}
where we have $\mathbf{W}^{\textit{proj}} \in \mathbb{R}^{C\times C}$ and $\hat{\mathbf{F}}_i$ is the enhanced output as in \secref{sec:overview}.

In \equref{eq:coord} - \equref{eq:residual}, each reference location attends to a small set of key sampling locations around the bilaterally symmetric location of the reference location, rather than just the symmetric location. The key sampling locations are automatically set in a learning way.
This ensures the receptive field when comparing the appearance of the left and right sides of the lungs.
In our experiments, we have observed that the learned coordinate shifts $\Delta \mathbf{p}^x_i$ usually fall within a range of 10\% of the width of the corresponding feature map, without any constraints. Consequently, for points situated far from the vertical centerline, they will search points on the symmetric side for feature aggregation. For points in proximity to the vertical centerline, they usually do not correspond to lung regions (see \figref{fig:vis}), and this does not impact our assumption of symmetric searching for CTD as we focus on detecting TB in lung regions.
In this paper, we empirically set $M=8$ and $K=4$.
Suppose $N=\frac{H}{2^{i+1}} \times \frac{W}{2^{i+1}}$, and the computational complexity can be expressed as $\mathcal{O}(NC^2)$. 
Thus, SymAttention is very efficient and flexible for application to the feature pyramid $\mathbf{F}$.

\subsection{TB Diagnosis Heads} \label{sec:heads}
In \secref{sec:overview}, we mention that there are two TB diagnosis heads: the TB infection area detection head and the CXR image classification head.
In this section, we introduce them in detail.
The detection head is based on RetinaNet \cite{lin2017focal}, a well-known one-stage object detector, consisting of two branches for bounding box classification and location regression.
In contrast to object detection for natural images, where each bounding box covers an object, each bounding box in our system is designed to cover a TB infection area.
The detection head learns to detect TB with \textit{two categories}: active TB and latent TB.
During clinical TB screening, most CXR cases do not have TB infections, making it easy for the detection head to introduce false positives.
To tackle this challenge, we add a CXR image classification head to conduct simultaneous CXR image classification and TB infection area detection.
We discard the detected TB areas if a CXR image is classified as non-TB. 
For simplicity, we stack several convolutions with pooling operations for the classification head.
There are five sequential convolution layers, each with 512 output channels and ReLU activation.
We then adopt global average pooling to obtain a global feature vector, followed by a fully connected layer with 3 output neurons for classification into \textit{three categories}: healthy, sick but non-TB, and TB.

\subsection{Two-stage Training Diagram} \label{sec:training}
Our SymFormer framework consists of two heads designed for CXR image classification and TB infection area detection, respectively.
In clinical settings, the number of non-TB cases significantly outweighs the number of TB cases.
Directly training the infection area detection head with non-TB cases would result in an excessive number of pure background supervisions.
Therefore, simultaneous training of the classification and detection heads is suboptimal.
Additionally, CXR images solely depict structures and organs in the chest, unlike natural images that have complex and diverse backgrounds.
If we first train the backbone network and the classification head, the backbone network for feature extraction would become overfitted and would not generalize well to infection area detection.
Furthermore, image classification mainly focuses on global features, while infection area detection requires fine-grained features for TB area localization.
As a result, training image classification first is also suboptimal.
Our proposed approach entails training the backbone network and the infection area detection head initially using only TB CXR images.
Then, we employ all CXR images to train the classification head by freezing the backbone network and the detection head.
This training strategy benefits from more specific bounding box annotations provided by the detection annotations, which mitigates the risk of overfitting. 
The fine-grained features learned through the infection area detection can also be easily transferred to CXR image classification.

\section{Experimental Setup}
In this section, we first elaborate on the implementation details for the proposed SymFormer in \secref{sec:impl}. Subsequently, we introduce several baseline models for CTD in \secref{sec:baselines} and discuss the evaluation metrics used for CTD in \secref{sec:metrics}.

\subsection{Implementation Details} \label{sec:impl}
Our implementation of SymFormer is done using PyTorch \cite{paszke2019pytorch} and the open-source \texttt{mmdetection} framework \cite{chen2019mmdetection}. 
The training of the first stage uses TB CXR images in the TBX11K \texttt{trainval} (\texttt{train} + \texttt{val}) set, while the training of the second stage not only uses all TBX11K \texttt{trainval} CXR images but also the random half of the MC \cite{jaeger2014two} and Shenzhen \cite{jaeger2014two} datasets as well as the training sets of the DA \cite{chauhan2014role} and DB \cite{chauhan2014role} datasets.
The other half of the MC \cite{jaeger2014two} and Shenzhen \cite{jaeger2014two} datasets as well as the test sets of the TBX11K, DA \cite{chauhan2014role} and DB \cite{chauhan2014role} datasets are used to evaluate the performance of CXR image classification.
We set the number of FPN feature channels, denoted as $C$, to 256, consistent with RetinaNet \cite{lin2017focal}. 
Other settings also follow those in RetinaNet.
For the training of the first stage, we use a batch size of 8 and train for 50 epochs for Deformable DETR-based models \cite{zhu2021deformable}, while 24 epochs for other models.
For the training of the second stage, we utilize a batch size of 8 and train for 12 epochs across all models.
We adopt the AdamW optimizer for Deformable DETR-based models and the SGD optimizer for other models.
To augment the data, we use random flipping. 
We resize both the CXR images used for training and testing to $512\times 512$.
All experiments are conducted using 2 TITAN XP GPUs.
Please refer to our code for more details.

\subsection{Baseline Models} \label{sec:baselines}
As discussed in \secref{sec:overview}, incorporating an image classification head can significantly reduce the false positives of detection in clinical TB screening. 
However, existing object detectors do not consider background images and often disregard images without bounding-box objects \cite{liu2016ssd,lin2017focal,tian2019fcos,ren2015faster,cheng2019bing}. 
Using these detectors directly for CTD leads to numerous false positives due to the large number of non-TB CXR images in clinical practice. 
To address this issue, we introduce a classification head to enable simultaneous CXR image classification and TB infection area detection, where the CXR image classification results are used to filter out the false positives of detection.

To achieve this, we reformulate several well-known object detectors, including SSD \cite{liu2016ssd}, RetinaNet \cite{lin2017focal}, Faster R-CNN \cite{ren2015faster}, FCOS \cite{tian2019fcos}, and Deformable DETR \cite{zhu2021deformable} for simultaneous CXR image classification and TB infection area detection. 
Specifically, we add the same image classification head as used in our SymFormer to these detectors, after the final layer of their backbone networks, \eg, \textit{conv5\_3} for VGGNet-16 \cite{simonyan2015very} and \textit{res5c} for ResNet-50 \cite{he2016deep}. 
The image classification head learns to classify CXR images into three categories: healthy, sick but non-TB, and TB, while the TB detection head learns to detect TB with two categories: active TB and latent TB. 
The training of existing detectors follows the two-stage training diagram described in \secref{sec:training}.

\begin{table*}[!t]
  \centering
  \setlength\tabcolsep{4.pt}
  \caption{\textbf{CXR image classification results (\%) in terms of accuracy, AUC, sensitivity, specificity, AP, and AR on the TBX11K test data.} The ``Backbone'' column indicates the specific backbone network used.}
  \label{tab:cls1}
  \begin{tabular}{l|c|c|c|c|c|c|c} \toprule
    Methods & Backbones & Accuracy & AUC (TB) & Sensitivity & Specificity & Ave. Prec. (AP) & Ave. Rec. (AR) \\ \midrule
    SSD \cite{liu2016ssd} & VGGNet-16 & 84.7 & 93.0 & 78.1 & 89.4 & 82.1 & 83.8 \\
    RetinaNet \cite{lin2017focal} & ResNet-50 w/ FPN & 87.4 & 91.8 & 81.6 & 89.8 & 84.8 & 86.8 \\
    Faster R-CNN \cite{ren2015faster} & ResNet-50 w/ FPN & 89.7 & 93.6 & 91.2 & 89.9 & 87.7 & 90.5 \\
    FCOS \cite{tian2019fcos} & ResNet-50 w/ FPN & 88.9 & 92.4 & 87.3 & 89.9 & 86.6 & 89.2 \\ 
    Deformable DETR \cite{zhu2021deformable} & ResNet-50 w/ FPN & 91.3 & 97.6 & 89.2 & 95.3 & 89.8 & 91.0 \\ \midrule
    SymFormer w/ Deformable DETR & ResNet-50 w/ FPN & 94.3 & 98.5 & 87.3 & \textbf{97.3} & 93.2 & 93.2 \\
    SymFormer w/ RetinaNet & ResNet-50 w/ FPN & 94.5 & 98.9 & 91.0 & 96.8 & 93.3 & 94.0 \\
    SymFormer w/ RetinaNet & P2T-Small w/ FPN & \textbf{94.6} & \textbf{99.1} & \textbf{92.1} & 96.7 & \textbf{93.4} & \textbf{94.2} \\
    \bottomrule
  \end{tabular}%
\end{table*}

\begin{table*}[!t]
  \centering
  \setlength\tabcolsep{1.8pt}
  \caption{\textbf{CXR image classification results (\%) in terms of the $F_1$ score and confusion matrix on the TBX11K test data, as well as the number of FLOPs, the number of parameters, and FPS of each model.} ``\#Total'' denotes the total number of test CXR images. We test FPS on a single TITAN XP GPU. For the ground truths, the ratio of positives (TP + FN) is 19.6\%, and the ratio of negatives (TN + FP) is 80.4\%.}
  \label{tab:cls2}
  \begin{tabular}{l|c|c|c|c|c|c|c|c|c} \toprule
    Methods & Backbones & \#FLOPs & \#Params & FPS & $F_1$ score $\uparrow$ & TP/\#Total $\uparrow$ & TN/\#Total $\uparrow$ & FP/\#Total $\downarrow$ & FN/\#Total $\downarrow$ \\ \midrule
    SSD \cite{liu2016ssd} & VGGNet-16 & 90.58 & 38.69 & 32.9 & 70.5 & 15.3 & 71.9 & 8.5 & 4.3 \\
    RetinaNet \cite{lin2017focal} & ResNet-50 w/ FPN & 55.41 & 48.97 & 35.3 & 73.1 & 16.0 & 72.2 & 8.2 & 3.6 \\
    Faster R-CNN \cite{ren2015faster} & ResNet-50 w/ FPN & 66.27 & 53.98 & 30.3 & 78.5 & 17.9 & 72.3 & 8.1 & 1.7 \\
    FCOS \cite{tian2019fcos} & ResNet-50 w/ FPN & 53.33 & 44.69 & 39.9 & 76.3 & 17.1 & 72.3 & 8.1 & 2.5 \\ 
    Deformable DETR \cite{zhu2021deformable} & ResNet-50 w/ FPN & 54.07 & 52.67 & 23.0 & 85.6 & 17.5 & 76.6 & 3.8 & 2.1 \\ \midrule
    SymFormer w/ Deformable DETR & ResNet-50 w/ FPN & 54.08 & 52.69 & 22.5 & 87.9 & 17.1 & \textbf{78.2} & \textbf{2.2} & 2.5 \\
    SymFormer w/ RetinaNet & ResNet-50 w/ FPN & 59.14 & 50.03 & 24.3 & 89.0 & 17.8 & 77.8 & 2.6 & 1.8 \\
    SymFormer w/ RetinaNet & P2T-Small w/ FPN & 55.46 & 45.10 & 17.9 & \textbf{89.6} & \textbf{18.1} & 77.7 & 2.7 & \textbf{1.5} \\
    \bottomrule
  \end{tabular}%
\end{table*}

\subsection{Evaluation Metrics} \label{sec:metrics}
\textbf{CXR image classification.}\quad
We continue by introducing the evaluation metrics for the CTD task.
In CXR image classification, the goal is to classify each CXR image into one of three categories: healthy, sick but non-TB, and TB.
To assess the classification results, we utilize the following evaluation metrics:
\begin{itemize}
\item Accuracy: This metric measures the percentage of CXR images that are correctly classified across all three categories.
\item Area Under Curve (AUC): AUC computes the area under the Receiver Operating Characteristic (ROC) curve. The ROC curve plots the true positive rate against the false positive rate for the TB class.
\item Sensitivity: Sensitivity quantifies the percentage of TB cases that are accurately identified as TB. It represents the recall for the TB class.
\item Specificity: Specificity determines the percentage of non-TB cases that are correctly identified as non-TB, encompassing both the healthy and sick but non-TB classes. It represents the recall for the non-TB class.
\item Average Precision (AP): AP calculates the precision for each class and takes the average across all classes. It provides an overall measure of precision.
\item Average Recall (AR): AR computes the recall for each class and averages the values across all classes. It provides an overall measure of recall.
\item Confusion matrix: The confusion matrix reports the number of true positives (TP), true negatives (TN), false positives (FP), and false negatives (FN). For a better view, we report the ratios of TP, TN, FP, and FN in relation to the total number of test CXR images.
\item $F_1$ score: This metric is the harmonic mean of precision and recall, thus symmetrically representing both in a single value. It can be calculated by 2TP/(2TP + FP + FN).
\end{itemize}
These metrics enable the evaluation of the CXR image classification quality from various perspectives.

\myPara{TB infection area detection.}
For the evaluation of TB detection, we utilize the average precision of the bounding box metric (AP$^\text{bb}$) proposed by the COCO benchmark \cite{lin2014microsoft}.
AP$^\text{bb}$ is widely used as the primary detection metric in the vision community \cite{lin2017focal,tian2019fcos,liu2022leveraging,wu2022p2t}.
The default AP$^\text{bb}$ is computed by averaging over IoU (intersection-over-union) thresholds ranging from 0.5 to 0.95 with a step size of 0.05.
Additionally, we report AP$_{\text{50}}^\text{bb}$, which represents AP$^\text{bb}$ at an IoU threshold of 0.5.
To provide insights into the detection performance for different types of TB, 
we present evaluation results separately for active TB and latent TB, excluding uncertain TB CXR images.
We also report category-agnostic TB detection results, where the TB categories are disregarded, to describe the detection of all TB areas. 
In this case, uncertain TB CXR images are included.
Furthermore, we introduce two evaluation modes: i) utilizing all CXR images in the TBX11K test set, and ii) considering only TB CXR images in the TBX11K test set.
By employing these metrics, we can comprehensively analyze the performance of CTD systems from various useful perspectives.

\begin{table*}[!t]
  \centering
  \setlength\tabcolsep{4.pt}
  \caption{\textbf{CXR image classification results (\%) in terms of accuracy, AUC, sensitivity, specificity, AP, and AR on the DA+DB test data \cite{chauhan2014role}.}}
  \label{tab:cls-da_db}
  \begin{tabular}{l|c|c|c|c|c|c|c} \toprule
    Methods & Backbones & Accuracy & AUC (TB) & Sensitivity & Specificity & Ave. Prec. (AP) & Ave. Rec. (AR) \\ \midrule
    SSD \cite{liu2016ssd} & VGGNet-16 & 51.0 & 53.8 & \textbf{100.0} & 1.9 & 75.3 & 51.0 \\
    RetinaNet \cite{lin2017focal} & ResNet-50 w/ FPN & 50.0 & 50.0 & \textbf{100.0} & 0.0 & 25.0 & 50.0 \\
    Faster R-CNN \cite{ren2015faster} & ResNet-50 w/ FPN & 50.0 & 51.9 & \textbf{100.0} & 0.0 & 25.0 & 50.0 \\
    FCOS \cite{tian2019fcos} & ResNet-50 w/ FPN & 50.0 & 52.1 & \textbf{100.0} & 0.0 & 25.0 & 50.0 \\ 
    Deformable DETR \cite{zhu2021deformable} & ResNet-50 w/ FPN & 68.6 & 69.7 & 84.3 & 52.9 & 70.7 & 68.6 \\ \midrule
    SymFormer w/ Deformable DETR & ResNet-50 w/ FPN & 82.4 & 78.4 & 86.3 & 78.4 & 82.6 & 82.4 \\
    SymFormer w/ RetinaNet & ResNet-50 w/ FPN & 78.4 & 74.7 & 90.2 & 66.7 & 80.1 & 78.4 \\
    SymFormer w/ RetinaNet & P2T-Small w/ FPN & \textbf{84.3} & \textbf{89.4} & 84.3 & \textbf{84.3} & \textbf{84.3} & \textbf{84.3} \\
    \bottomrule
  \end{tabular}%
\end{table*}

\begin{table*}[!t]
  \centering
  \setlength\tabcolsep{4.pt}
  \caption{\textbf{CXR image classification results (\%) in terms of accuracy, AUC, sensitivity, specificity, AP, and AR on the MC+Shenzhen test data \cite{jaeger2014two}.}}
  \label{tab:cls-mc_sz}
  \begin{tabular}{l|c|c|c|c|c|c|c} \toprule
    Methods & Backbones & Accuracy & AUC (TB) & Sensitivity & Specificity & Ave. Prec. (AP) & Ave. Rec. (AR) \\ \midrule
    SSD \cite{liu2016ssd} & VGGNet-16 & 50.8 & 50.4 & \textbf{100.0} & 3.4 & 74.9 & 51.7 \\
    RetinaNet \cite{lin2017focal} & ResNet-50 w/ FPN & 49.3 & 49.7 & \textbf{100.0} & 0.5 & 74.6 & 50.3 \\
    Faster R-CNN \cite{ren2015faster} & ResNet-50 w/ FPN & 49.0 & 49.5 & \textbf{100.0} & 0.0 & 24.5 & 50.0 \\
    FCOS \cite{tian2019fcos} & ResNet-50 w/ FPN & 48.8 & 49.0 & 99.5 & 0.0 & 24.4 & 49.7 \\ 
    Deformable DETR \cite{zhu2021deformable} & ResNet-50 w/ FPN & 81.3 & 83.5 & 92.9 & 70.1 & 83.0 & 81.5 \\ \midrule
    SymFormer w/ Deformable DETR & ResNet-50 w/ FPN & 82.0 & 84.7 & 89.3 & 75.0 & 82.7 & 82.1 \\
    SymFormer w/ RetinaNet & ResNet-50 w/ FPN & 82.8 & 86.3 & 91.8 & 74.0 & 83.8 & 82.9 \\
    SymFormer w/ RetinaNet & P2T-Small w/ FPN & \textbf{85.8} & \textbf{87.4} & 93.4 & \textbf{78.4} & \textbf{86.6} & \textbf{85.9} \\
    \bottomrule
  \end{tabular}%
  \vspace{-.5mm}
\end{table*}

\begin{table*}[!t]
  \centering
  \setlength\tabcolsep{8.pt}
  \caption{\textbf{TB infection area detection results (\%) on our TBX11K test set.} The ``Test Data'' column specifies whether the evaluation was performed using all CXR images in the test set or only TB CXR images in the test set. The ``Backbone'' column indicates the specific backbone network used.}
  \label{tab:detection}
  \begin{tabular}{l|c|c|cc|cc|cc} \toprule
    \multirow{2}{*}{Methods} & \multirow{2}{*}{Test Data} & \multirow{2}{*}{Backbones} & \multicolumn{2}{c|}{Category-agnostic TB} & \multicolumn{2}{c|}{Active TB} & \multicolumn{2}{c}{Latent TB} \\ \cmidrule{4-9}
    & & & AP$_{\text{50}}^\text{bb}$ & AP$^\text{bb}$ & AP$_{\text{50}}^\text{bb}$ & AP$^\text{bb}$ & AP$_{\text{50}}^\text{bb}$ & AP$^\text{bb}$ \\ \midrule
    SSD \cite{liu2016ssd} & \multirow{8}{*}{ALL} & VGGNet-16 & 52.3 & 22.6 & 50.5 & 22.8 & 8.1 & 3.2 \\
    RetinaNet \cite{lin2017focal} & & ResNet-50 w/ FPN & 52.1 & 22.2 & 45.4 & 19.6 & 6.2 & 2.4 \\
    Faster R-CNN \cite{ren2015faster} & & ResNet-50 w/ FPN & 57.3 & 22.7 & 53.3 & 21.9 & 9.6 & 2.9  \\
    FCOS \cite{tian2019fcos} & & ResNet-50 w/ FPN & 46.6 &  18.9 & 40.3 & 16.8 & 6.2 & 2.1  \\
    Deformable DETR \cite{zhu2021deformable} & & ResNet-50 w/ FPN & 51.7 & 22.0 & 48.9 & 21.2 & 7.1& 1.9 \\
    SymFormer w/ Deformable DETR & & ResNet-50 w/ FPN & 57.0 & 23.3 & 52.1 & 22.7 & 7.1 & 2.0 \\
    SymFormer w/ RetinaNet & & ResNet-50 w/ FPN & 68.0 & 29.5 & 62.0 & \textbf{27.3} & \textbf{13.3} & \textbf{4.4} \\ 
    SymFormer w/ RetinaNet & & P2T-Small w/ FPN & \textbf{70.4} & \textbf{30.0} & \textbf{63.6} & 26.9 & 11.4 & 4.3 \\
    \midrule
    SSD \cite{liu2016ssd} & \multirow{8}{*}{Only TB} & VGGNet-16 & 68.3 & 28.7 & 63.7 & 28.0 & 10.7 & 4.0   \\
    RetinaNet \cite{lin2017focal} & & ResNet-50 w/ FPN & 69.4 & 28.3 & 61.5 & 25.3 & 10.2 & 4.1   \\
    Faster R-CNN \cite{ren2015faster} & & ResNet-50 w/ FPN & 63.4 & 24.6 & 58.7 & 23.7 & 9.6 & 2.8 \\
    FCOS \cite{tian2019fcos} & & ResNet-50 w/ FPN & 56.3 & 22.5 & 47.9 & 19.8 & 7.4 & 2.4 \\
    Deformable DETR \cite{zhu2021deformable} & & ResNet-50 w/ FPN & 57.4 & 24.2 & 54.5 & 23.5 & 7.6 & 2.3 \\
    SymFormer w/ Deformable DETR & & ResNet-50 w/ FPN & 60.8 & 24.5 & 55.2 & 23.8 & 9.2 & 2.6 \\
    SymFormer w/ RetinaNet & & ResNet-50 w/ FPN & 73.4 & 31.5 & 67.1 & \textbf{29.2} & \textbf{14.7} & \textbf{4.8} \\ 
    SymFormer w/ RetinaNet & & P2T-Small w/ FPN & \textbf{75.7} & \textbf{32.1} & \textbf{68.9} & 28.9 & 13.0 & 4.7 \\ \bottomrule
  \end{tabular}%
\end{table*}

\section{Experimental Results}
In this section, we present the results for CXR image classification in \secref{sec:cls}, followed by the results for TB infection area detection in \secref{sec:detection}. 
Subsequently, we visualize detection results and the learned deep features in \secref{sec:vis}.
Lastly, we conduct ablation studies in \secref{sec:ablation} to gain a better understanding of the proposed SymFormer.

\subsection{CXR Image Classification} \label{sec:cls}
We summarize the evaluation results for CXR image classification on the TBX11K test set in \tabref{tab:cls1} and \tabref{tab:cls2}.
All methods adopt pretraining models from ImageNet \cite{deng2009imagenet} for initialization.
We report the results of the proposed SymFormer integrated with RetinaNet \cite{lin2017focal} and Deformable DETR \cite{zhu2021deformable} as the base methods.
As can be observed from both \tabref{tab:cls1} and \tabref{tab:cls2}, incorporating SymFormer into RetinaNet \cite{lin2017focal} and Deformable DETR \cite{zhu2021deformable} leads to significant performance improvements for RetinaNet and Deformable DETR, respectively.
SymFormer with Deformable DETR achieves a specificity of 97.3\%, indicating that 2.7 out of 100 non-TB CXR images will be misclassified as TB.
The \textit{default model} we employ is SymFormer with RetinaNet, which exhibits slightly lower performance than SymFormer with Deformable DETR but outperforms the latter by a significant margin in object detection, as demonstrated in \secref{sec:detection}.
Furthermore, in terms of accuracy, all methods greatly outperform radiologists who achieve an accuracy of 84.8\% as in \secref{sec:human}. 
This emphasizes the promising potential of deep-learning-based CTD as a research field.

In \tabref{tab:cls1}, we observe that the difference between SymFormer and the baseline models in terms of sensitivity is much smaller than the difference in terms of specificity. Specially, Faster R-CNN \cite{ren2015faster} achieves an impressively high sensitivity rate of 91.2\%, but it lags significantly behind SymFormer in other performance metrics. To explain this phenomenon, we refer to \tabref{tab:cls2} and discover that the baseline models tend to make more positive predictions (TP + FP) and fewer negative predictions (TN + FN). In simpler terms, the baseline models are inclined to classify a test CXR image as positive, potentially due to their limited ability to learn high-quality TB-related features. When we evaluate all models on other public datasets without retraining, as shown in \tabref{tab:cls-da_db} and \tabref{tab:cls-mc_sz}, we can see that baseline models even achieve a sensitivity rate of 100.0\% and a specificity rate of 0. This further confirms our hypothesis that baseline models tend to classify CXR images as positive. Considering this perspective, the $F_1$ score in \tabref{tab:cls2} provides a better representation of a model's overall performance, as it symmetrically combines precision and recall.

In \tabref{tab:cls2}, we also report the number of Floating-Point Operations (FLOPs), the number of parameters, and Frames Per Second (FPS) for each model. From the comparison between the original and SymFormer-enhanced Deformable DETR \cite{zhu2021deformable}, we can see that SymFormer exhibits similar FLOPs, parameters, and running speed as Deformable DETR. This is straightforward as SymFormer only introduces negligible computations to Deformable DETR. When integrated with RetinaNet \cite{lin2017focal} using the ResNet-50 \cite{he2016deep} backbone, SymFormer achieves 24.3 fps, making it a practical choice for deployment in real-world scenarios. If additional computational resources are available, SymFormer can also utilize P2T-Small \cite{wu2022p2t} as the backbone, offering enhanced diagnostic performance and a speed of 17.9 fps.

\newcommand{\AddImg}[1]{\includegraphics[width=.19\linewidth]{#1}}

\begin{figure*}[!t]
  \centering
  \begin{sideways} \footnotesize \hspace{0.45in} ALL \end{sideways}
  \AddImg{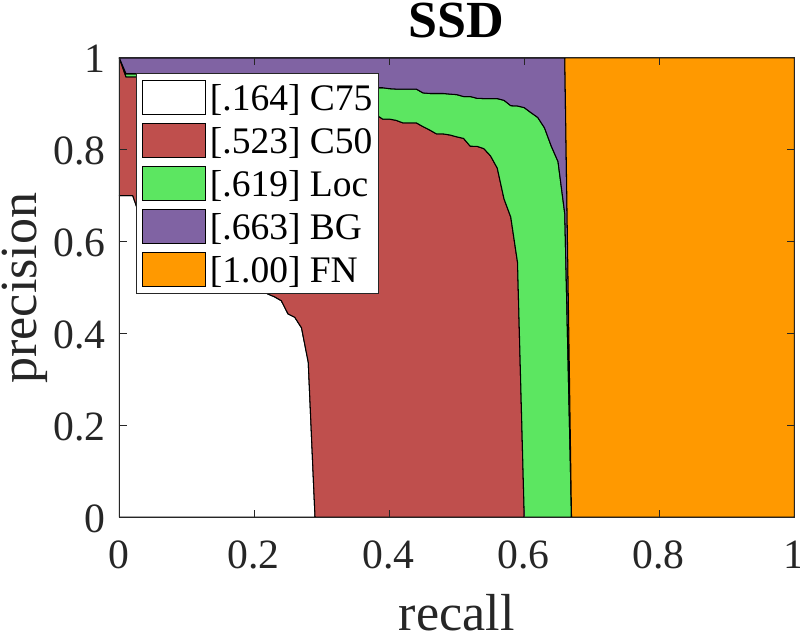} \hfill \AddImg{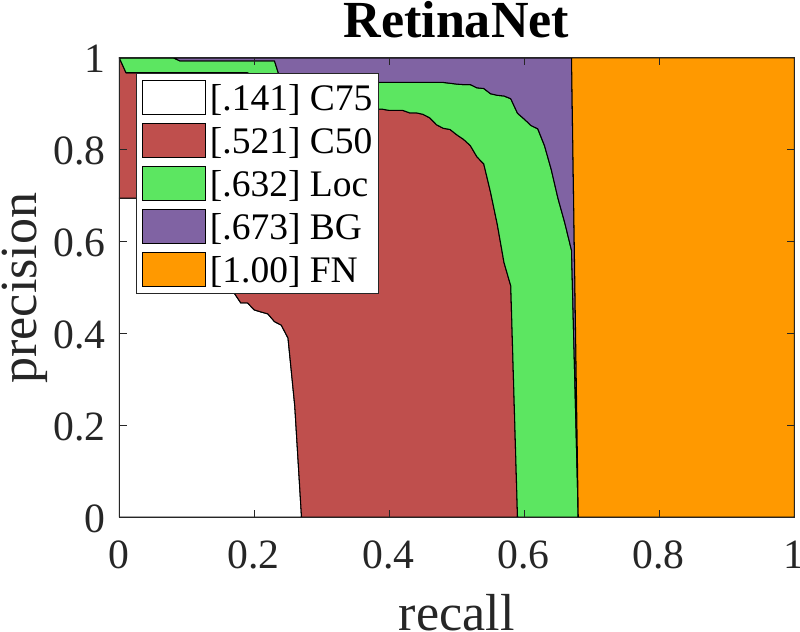} \hfill
  \AddImg{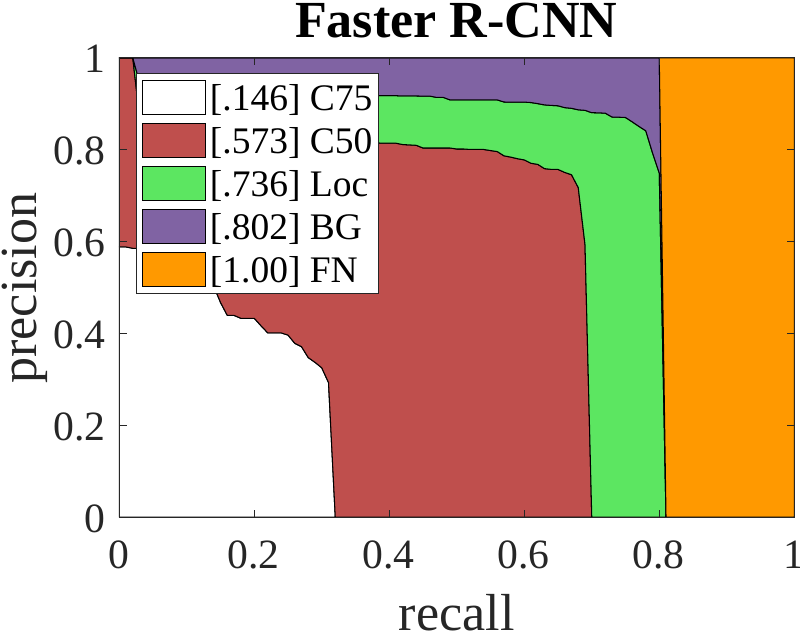} \hfill \AddImg{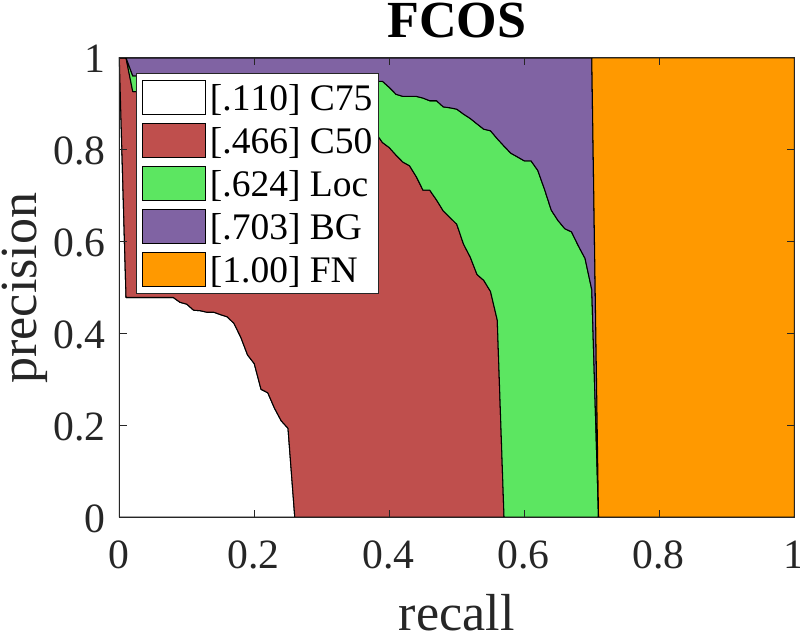} \hfill
  \AddImg{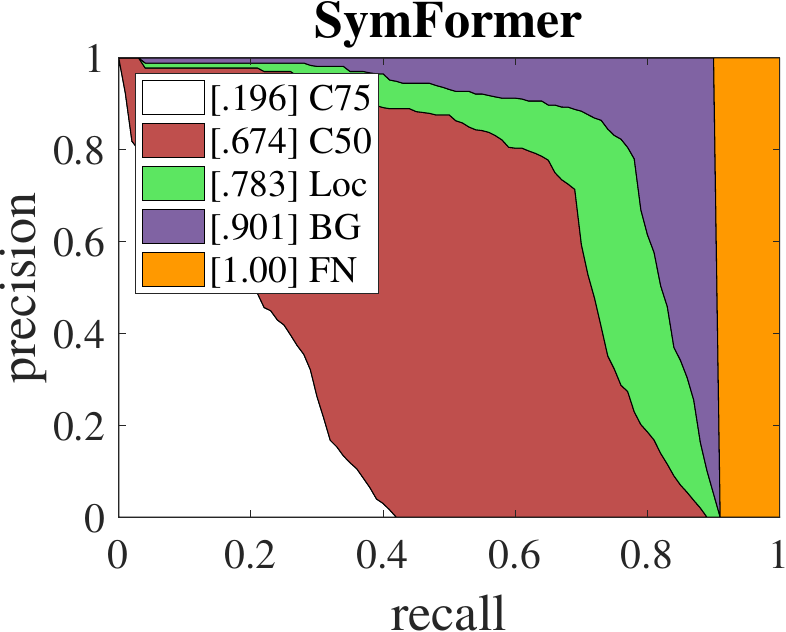}  \\ \vspace{0.05in}
  \begin{sideways} \footnotesize \hspace{0.35in} Only TB \end{sideways}
  \AddImg{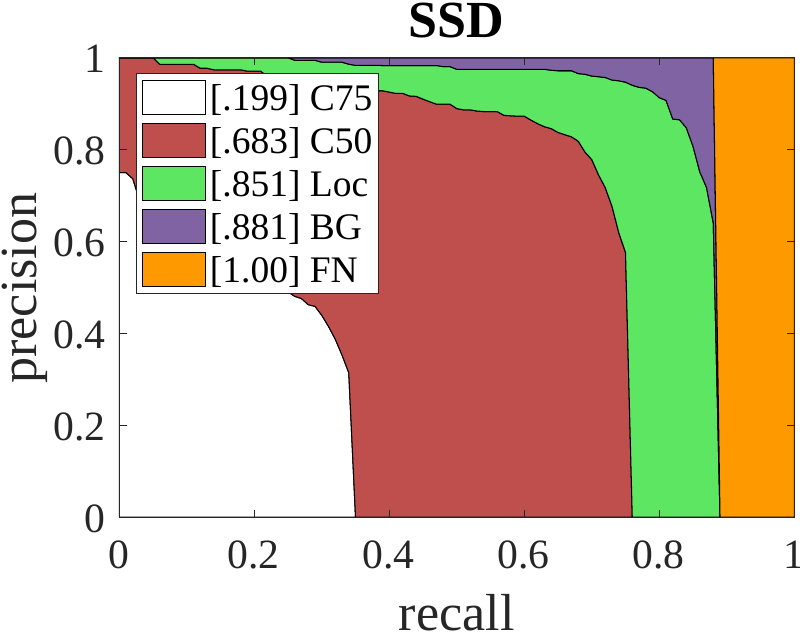} \hfill \AddImg{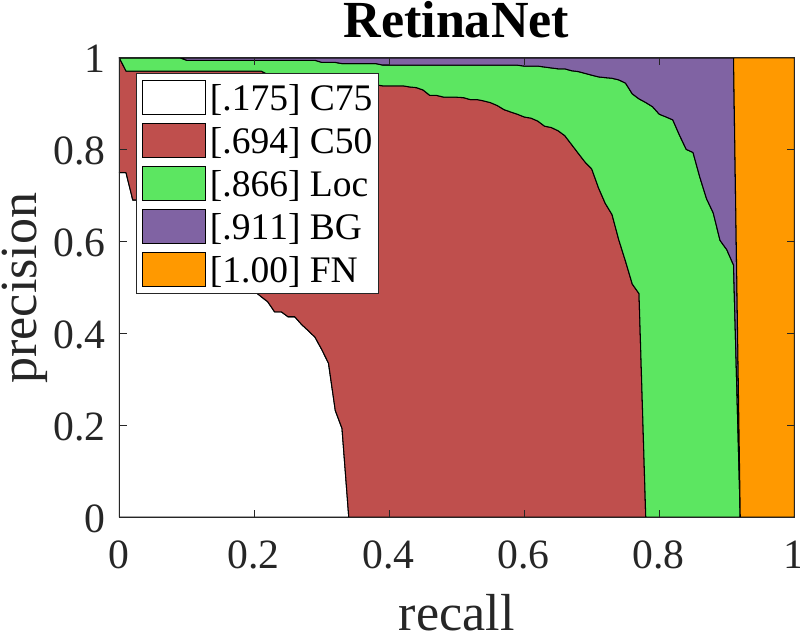} \hfill
  \AddImg{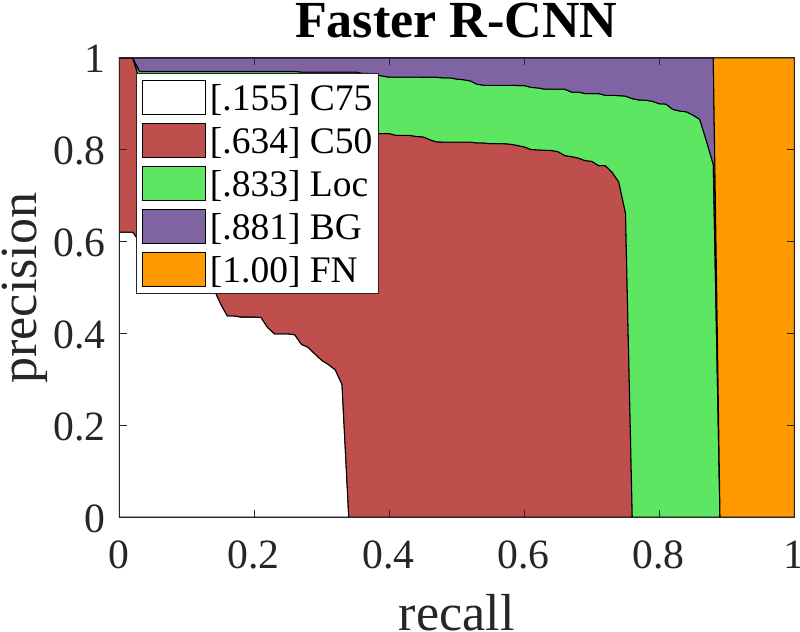} \hfill \AddImg{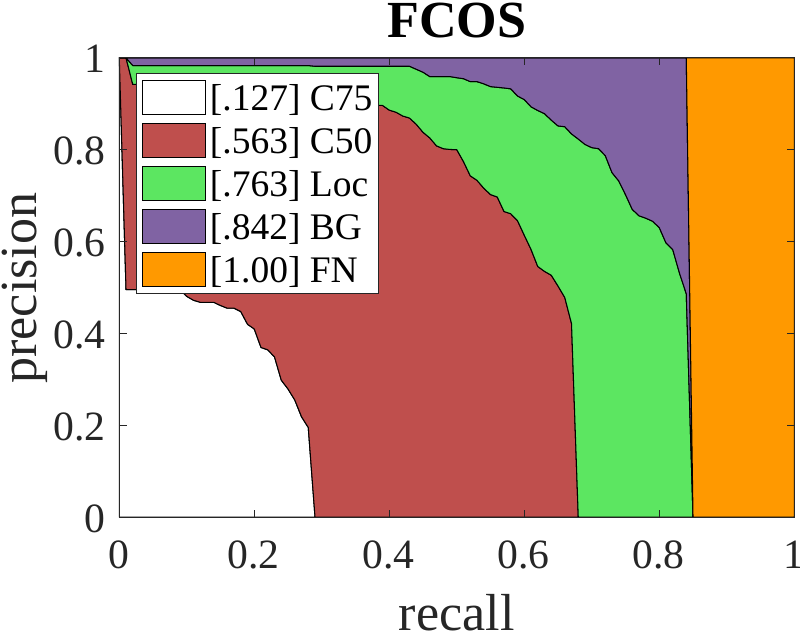}  \hfill
  \AddImg{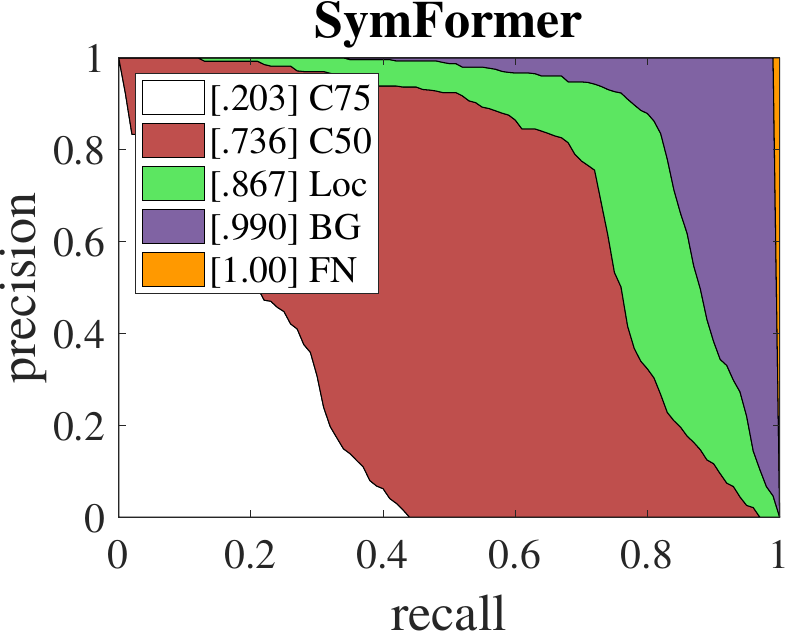}
  \caption{\textbf{Error analyses of category-agnostic TB area detection using baseline models and SymFormer w/ RetinaNet.} The first row is evaluated using all CXR images, while the second row only uses TB CXR images. C50/C75: PR curves under IoU thresholds of 0.5/0.75. Loc: the PR curve under the IoU threshold of 0.1. BG: removing background false positives. FN: removing other errors caused by undetected targets (false negatives). SymFormer largely outperforms other methods in all metrics, \eg, obtaining a remarkable 99\% BG score when only using TB CXR images.}
  \label{fig:err_ana}
\end{figure*}

In addition, we assess CXR image classification on public datasets using the above-trained models without fine-tuning. The results are presented in \tabref{tab:cls-da_db} for DA+DB test data \cite{chauhan2014role} and \tabref{tab:cls-mc_sz} for MC+Shenzhen test data \cite{jaeger2014two}. Notably, SSD \cite{liu2016ssd}, RetinaNet \cite{lin2017focal}, Faster R-CNN \cite{ren2015faster}, and FCOS \cite{tian2019fcos} achieve a sensitivity rate of approximately 100.0\% and a specificity rate of around 0. As discussed earlier, this suggests that baseline models struggle to learn robust feature representations for CTD, often misclassifying test CXR images as TB cases. Deformable DETR \cite{zhu2021deformable} demonstrates a degree of generalization to public datasets but falls short in comparison to the proposed SymFormer. The strong performance of SymFormer underscores its exceptional generalization capability.

\subsection{TB Infection Area Detection} \label{sec:detection}
We proceed by presenting the results for TB infection area detection.
As discussed in \secref{sec:metrics}, we report the performance for both the entire TBX11K test set and a subset consisting only of TB CXR images.
Evaluating the performance using only TB CXR images allows for precise detection analysis since non-TB CXR images do not contain target TB infection areas. 
Conversely, evaluating using all CXR images incorporates the influence of false positives in non-TB CXR images.
To ensure accurate evaluation using all CXR images, we \textit{discard} all predicted boxes in CXR images that are classified as non-TB by the CXR image classification head. 
However, it is important to note that this filtering process is not applicable when evaluating using only TB CXR images.

The results for TB infection area detection are presented in \tabref{tab:detection}. 
It is evident that both SymFormer with Deformable DETR and SymFormer with RetinaNet demonstrate significant improvements over their respective base methods, Deformable DETR \cite{zhu2021deformable} and RetinaNet \cite{lin2017focal}.
Interestingly, SymFormer with RetinaNet outperforms SymFormer with Deformable DETR by a considerable margin, indicating that SymFormer is better suited for integration with the RetinaNet framework. 
As a result, we select SymFormer with RetinaNet as our default model for CTD.
It is worth noting that all methods struggle with accurately detecting latent TB areas. 
However, the evaluation results for category-agnostic TB are better than those for active TB, indicating that many latent TB targets are correctly located but mistakenly classified as active TB. 
We attribute this to the limited number of latent TB CXR images in the TBX11K dataset, where only 212 CXR images depict latent TB compared to 924 CXR images depicting active TB. 
Therefore, future research should address this data imbalance issue and focus on improving the detection of latent TB areas.
Furthermore, we observe that the performance in terms of AP$_{\text{50}}^\text{bb}$ is generally superior to that of AP$^\text{bb}$. 
This suggests that while detection models are capable of identifying the target regions, their localization accuracy is often not very precise. 
We argue that locating TB bounding box regions differs significantly from locating regions of natural objects. 
Even experienced radiologists find it challenging to precisely pinpoint TB regions. 
Consequently, AP$_{\text{50}}^\text{bb}$ is more crucial than AP$^\text{bb}$ since predicted boxes with an IoU of 0.5 with target TB areas are sufficient to assist radiologists in identifying TB infection areas.

\begin{table}[!t]
  \centering
  \setlength\tabcolsep{3.pt}
  \caption{\textbf{Ablation study for TB infection area detection on our TBX11K validation set.} We only use TB CXR images to evaluate category-agnostic TB area detection. The ``Symmetry of SPE'' column indicates whether SPE transfers the right side of positional encoding to the left side, or vice versa. APE: absolute positional encoding; RPE: relative positional encoding.}
  \label{tab:ablation}
  \begin{tabular}{l|c|c|c|c} \toprule
    Attention & Positional Encoding & Symmetry of SPE & AP$_{\text{50}}^\text{bb}$ & AP$^\text{bb}$
    \\ \midrule
    No & No & - & 72.7 & 31.0 \\
    Vanilla & APE & - & 73.4 & 30.6 \\
    Vanilla & RPE & - & 72.7 & 29.7 \\
    Vanilla & SPE w/o STN & left $\to$ right & 74.0 & 30.5 \\
    Vanilla & SPE w/o STN & right $\to$ left & 74.3 & 30.8 \\
    Vanilla & SPE & left $\to$ right & 75.1 & 30.4 \\
    Vanilla & SPE & right $\to$ left & 75.7 & 29.6 \\
    SymAttention & APE & - & 74.9 & 30.0 \\
    SymAttention & RPE & - & 73.6 & 29.1 \\
    SymAttention & SPE w/o STN & left $\to$ right & 75.3 & 31.4 \\
    SymAttention & SPE w/o STN & right $\to$ left & 75.5 & 30.7 \\
    SymAttention & SPE & left $\to$ right & 76.3 & 30.9 \\
    \rowcolor{mygray}
    SymAttention & SPE & right $\to$ left & \textbf{76.6} & \textbf{31.7} \\ \bottomrule
  \end{tabular}
\end{table}

\renewcommand{\AddImg}[1]{\includegraphics[width=.16\linewidth]{Visualization/#1}}

\begin{figure*}[!t]
  \centering  
  \AddImg{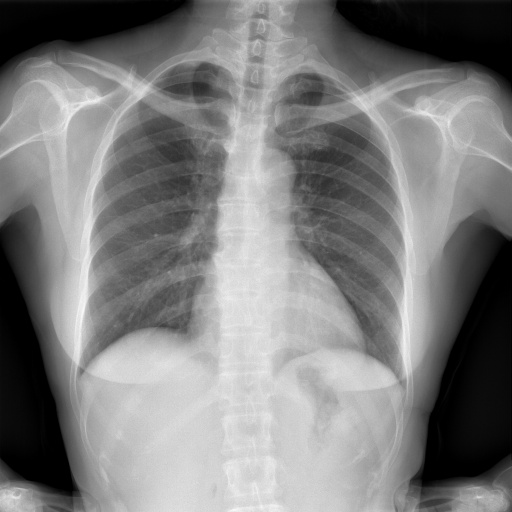} \hfill \AddImg{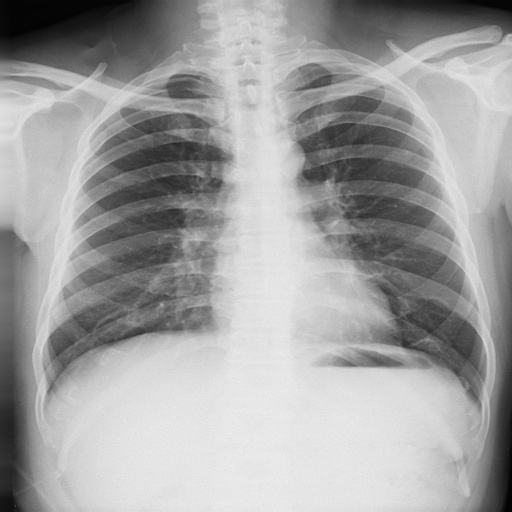} \hfill
  \AddImg{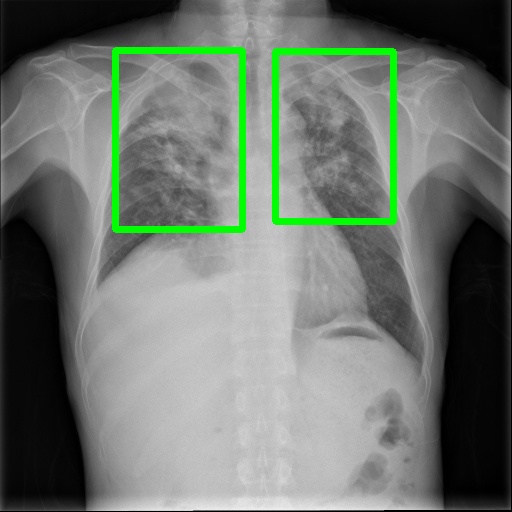} \hfill \AddImg{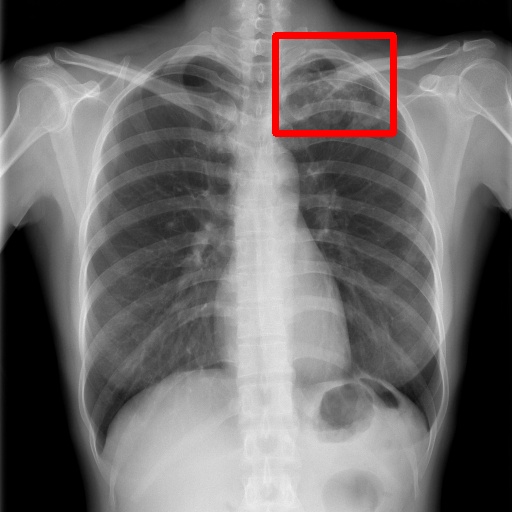} \hfill
  \AddImg{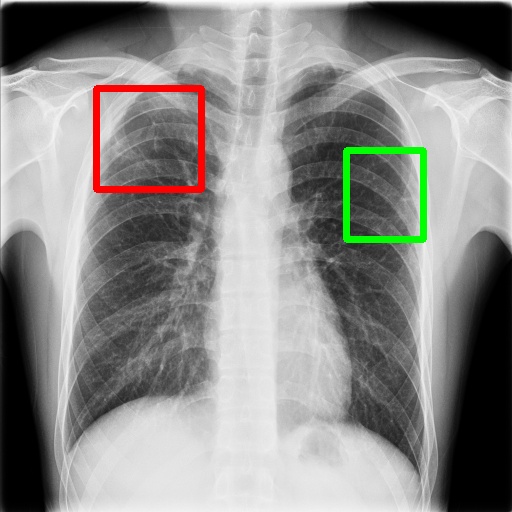} \hfill \AddImg{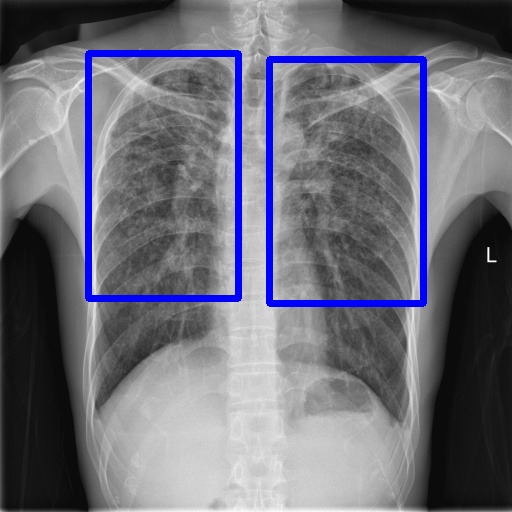} 
  \\ \vspace{0.01in}
  \AddImg{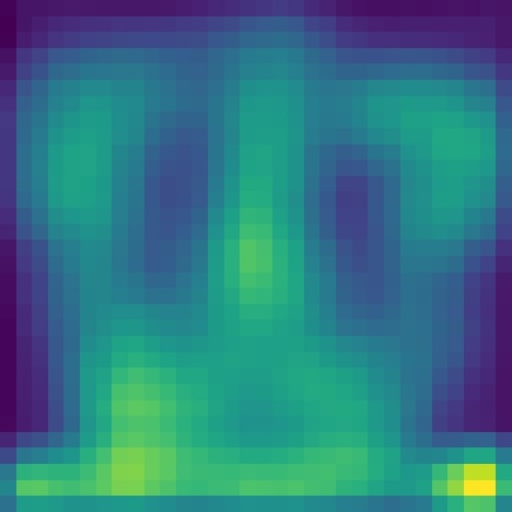} \hfill \AddImg{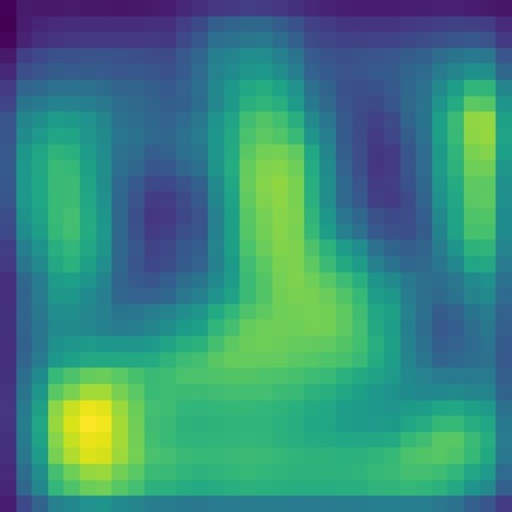} \hfill
  \AddImg{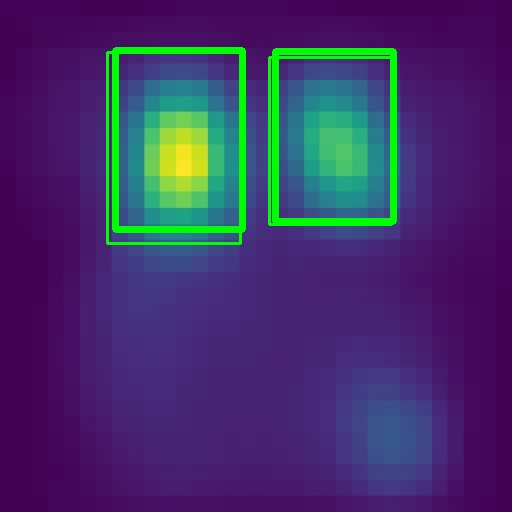} \hfill \AddImg{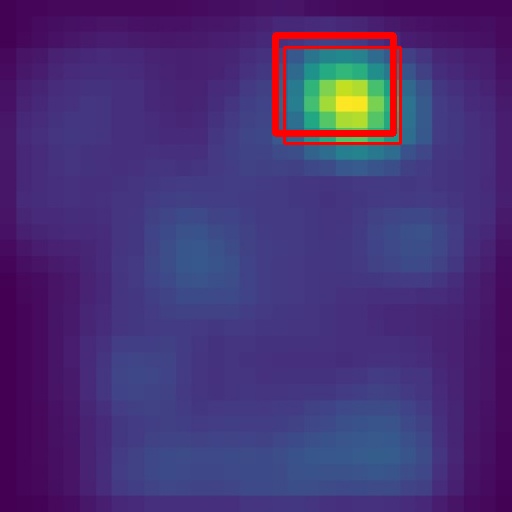} \hfill
  \AddImg{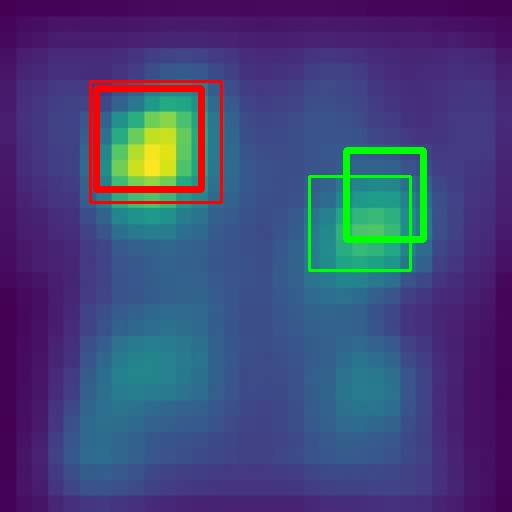} \hfill \AddImg{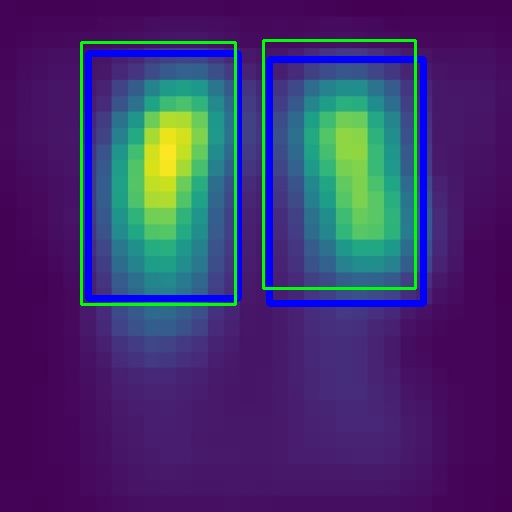}
  \\ \vspace{-0.06in}
  \leftline{\footnotesize \hspace{0.35in} Healthy \hspace{0.54in} Sick but Non-TB \hspace{0.50in} Active TB \hspace{0.64in} Latent TB \hspace{0.43in} Active \& Latent TB \hspace{0.35in} Uncertain TB}
  \caption{\textbf{Visualization of the learned deep features from CXR images using SymFormer w/ RetinaNet.} We randomly select CXR images from the TBX11K test set, and for each class mentioned in \tabref{tab:split}, we provide one example. In each example, the infection areas of active TB, latent TB, and uncertain TB are indicated by boxes colored in \textcolor{green}{green}, \textcolor{red}{red}, and \textcolor{blue}{blue}, respectively. The ground-truth boxes are displayed with thick lines, while the detected boxes are shown with thin lines.}
  \label{fig:vis}
\end{figure*}

\begin{table*}[!t]
  \centering
  \setlength\tabcolsep{4.pt}
  \caption{\textbf{Evaluation results (\%) of the 4-fold cross-validation.} The used model is our SymFormer w/ RetinaNet using ResNet-50 w/ FPN as the backbone. We split the TBX11K \texttt{trainval} set into 4 folds, each of which has a similar class distribution. ``\#Total'' denotes the total number of test CXR images in each fold. We only use TB CXR images to evaluate category-agnostic TB area detection.}
  \label{tab:cross-val}
  \begin{tabular}{c||c|c|c|c|c|c||c|c|c|c|c||c|c} \toprule
    Fold & Accuracy & AUC & Sensitivity & Specificity & AP & AR & $F_1$ score & TP/\#Total & TN/\#Total & FP/\#Total & FN/\#Total & AP$_{\text{50}}^\text{bb}$ & AP$^\text{bb}$ \\ \midrule
    1 & 94.7 & 98.9 & 92.2 & 97.5 & 91.1 & 94.6 & 87.6 & 11.0 & 85.8 & 2.2 & 0.9 & 74.7 & 31.7 \\
    2 & 95.2 & 98.9 & 92.6 & 97.6 & 91.7 & 95.0 & 88.1 & 11.1 & 86.0 & 2.1 & 0.9 & 75.2 & 30.4 \\
    3 & 94.6 & 99.1 & 92.9 & 96.9 & 90.5 & 94.6 & 86.4 & 11.1 & 85.3 & 2.7 & 0.8 & 74.1 & 29.5 \\
    4 & 95.1 & 99.3 & 92.9 & 97.3 & 91.2 & 94.7 & 87.4 & 11.1 & 85.7 & 2.4 & 0.8 & 75.4 & 33.3 \\
    \bottomrule
  \end{tabular}%
\end{table*}

In \figref{fig:err_ana}, we present the precision-recall (PR) curves for the detection error analyses, focusing on category-agnostic TB detection. 
It is evident that all methods exhibit substantial improvements when transitioning from an IoU threshold of 0.75 to 0.5. 
This indicates that the performance of all methods is particularly challenged at higher IoU thresholds due to their limited object localization capabilities.
Comparing the results obtained using all CXR images with those using only TB CXR images, we observe that the region labeled as ``FN'' (false negatives) is larger when evaluating using all CXR images. 
This suggests that the filtering process based on image classification disregards many correctly detected TB areas, despite its effectiveness in improving overall detection performance.
Importantly, the ``FN'' region for SymFormer is significantly smaller than that of other methods, highlighting its superior ability to detect fewer false negatives. 
Regardless of whether all CXR images or only TB CXR images are utilized, SymFormer consistently exhibits higher PR curves for IoU thresholds of 0.75, 0.5, and 0.1.
By considering the results of both image classification and TB infection area detection, we can confidently conclude that the proposed SymFormer achieves state-of-the-art performance and serves as a strong baseline for future research in the field of CTD.

\subsection{Ablation Study} \label{sec:ablation}
In this part, we first carry out ablation studies to investigate the effectiveness of the proposed modules.
Specifically, we train the models using the training set of our TBX11K dataset and evaluate them on the validation set.
The results are presented in \tabref{tab:ablation}.
The baseline model is RetinaNet \cite{lin2017focal}, which corresponds to the first model in \tabref{tab:ablation} and does not incorporate any attention or positional encoding.
The term ``vanilla attention'' refers to the deformable attention employed in Deformable DETR \cite{zhu2021deformable}.
We utilize well-established implementations for both absolute positional encoding \cite{vaswani2017attention,dosovitskiy2021image} (as described in \equref{eq:pos_enc}) and relative positional encoding \cite{kirillov2023segment}.
As specified in \equref{eq:spe}, the default version of SPE transfers the right side of the positional encoding to the left side.
Here, we also evaluate the performance when transferring the left side to the right side.

\renewcommand{\AddImg}[1]{%
\includegraphics[width=.16\linewidth]{Samples/#1_ssd} &
\includegraphics[width=.16\linewidth]{Samples/#1_retina} &
\includegraphics[width=.16\linewidth]{Samples/#1_faster} &
\includegraphics[width=.16\linewidth]{Samples/#1_fcos} &
\includegraphics[width=.16\linewidth]{Samples/#1_def} &
\includegraphics[width=.16\linewidth]{Samples/#1_sym_retina}}

\begin{figure*}[!t]
  \centering
  \scriptsize
  \setlength\tabcolsep{1.5pt}
  \begin{tabular}{cccccc}
    \AddImg{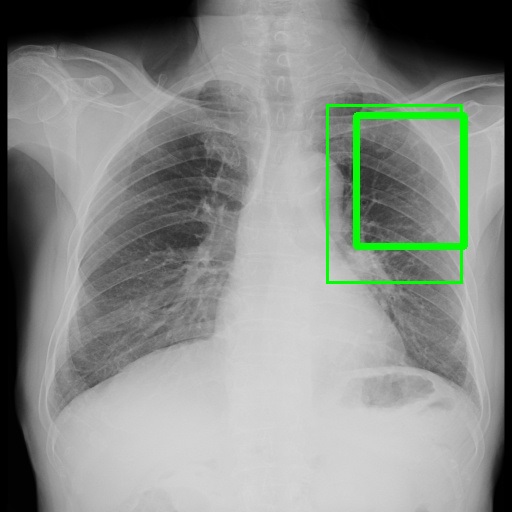} \\ \AddImg{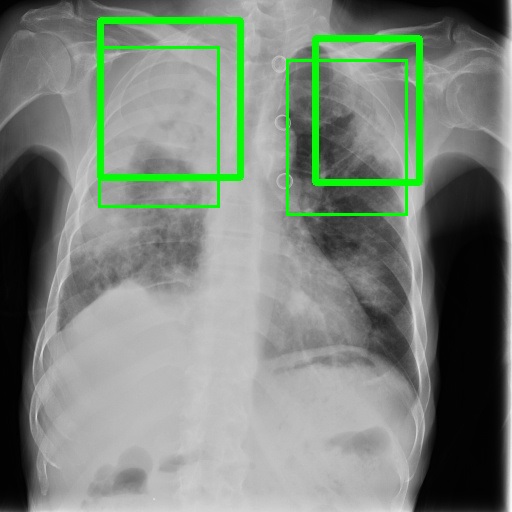} \\
    \AddImg{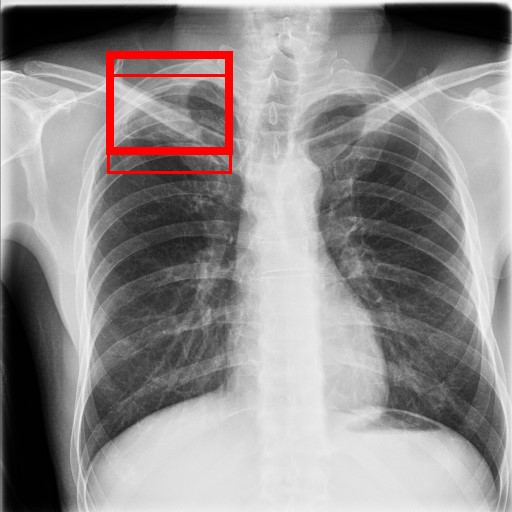} \\ \AddImg{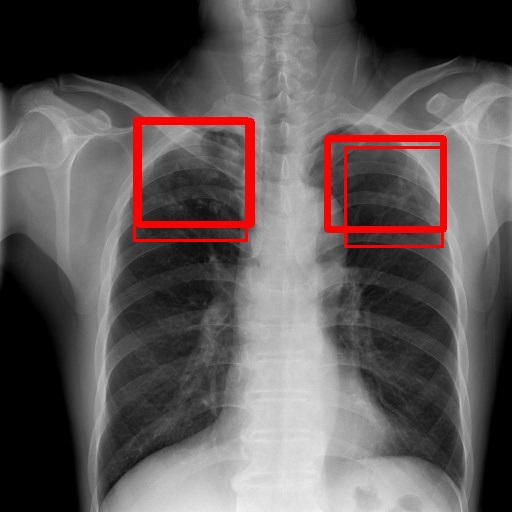} \\
    SSD \cite{liu2016ssd} & RetinaNet \cite{lin2017focal} & Faster R-CNN \cite{ren2015faster} & FCOS \cite{tian2019fcos} & Deformable DETR \cite{zhu2021deformable} & SymFormer w/ RetinaNet \\
  \end{tabular}
  \caption{\textbf{Qualitative comparison between the proposed SymFormer and baseline methods.} In each example, the infection areas of active TB and latent TB are indicated by \textcolor{green}{green} boxes and \textcolor{red}{red} boxes, respectively. The ground-truth boxes are displayed with thick lines, while the detected boxes are shown with thin lines. For all examples, SymFormer can detect all TB infection areas with true categories.}
  \label{fig:vis-comp}
\end{figure*}

Based on the discussions in \secref{sec:detection}, the AP$_{\text{50}}^\text{bb}$ metric is deemed sufficient for measuring the effectiveness of a model in assisting radiologists with identifying TB infection areas.
As evident from \tabref{tab:ablation}, relative positional encoding achieves inferior performance compared to absolute positional encoding, leading us to construct our SPE using absolute positional encoding.
Besides, the addition of absolute positional encoding and any form of attention to RetinaNet \cite{lin2017focal} yields significant improvements in detection performance. 
Furthermore, across all types of positional encoding, our proposed SymAttention consistently outperforms deformable attention, showcasing its superiority in learning distinctive representations for CTD.
Notably, even without STN, the proposed SPE consistently achieves superior performance compared to both absolute positional encoding and relative positional encoding. 
The inclusion of STN further enhances the performance of SPE, confirming its effectiveness.
Therefore, our investigation into symmetric abnormality search in CTD has yielded successful results.
In addition, we can observe that, for the symmetry of SPE, the transfer of positional encoding from right to left, as opposed to left to right, slightly outperforms.
Thus, we transfer the positional encoding from right to left by default.

To demonstrate the robustness of the proposed SymFormer, we perform a 4-fold cross-validation. We partition the TBX11K \texttt{trainval} data into four folds, ensuring that each fold maintains a similar class distribution. During each trial, we train SymFormer on three of these folds and evaluate its performance on the remaining fold. The evaluation results of the four trials are displayed in \tabref{tab:cross-val}. As can be observed, the results across these different trials are remarkably consistent, affirming the robustness of SymFormer.

\subsection{Visualization} \label{sec:vis}
To gain insights into the learning process of deep neural networks on CXR images, we visualize the feature map of SymFormer w/ RetinaNet at a scale of $1/32$.
To achieve this, we employ principal component analysis (PCA) to reduce the channels of the feature map to a single channel.
The resulting single-channel map is then converted into a heat map for visualization purposes.
The visualization of the learned features, along with the corresponding detection results, are presented in \figref{fig:vis}.
Upon analysis, we observe that the visualization of healthy cases exhibits irregular feature patterns, indicating the absence of significant abnormalities.
In contrast, the visualization of sick but non-TB cases displayed some discernible highlights, potentially representing the presence of lesions.
For TB cases, the highlights in the visualization map align well with the annotated TB infection areas, thereby indicating the effectiveness of the proposed SymFormer in learning deep features for TB area detection.
Furthermore, in \figref{fig:vis-comp}, we offer qualitative comparisons between the proposed SymFormer and the baseline models for TB infection area detection. As evident, SymFormer consistently delivers superior qualitative detection results.

\section{Conclusion}
Early diagnosis plays a crucial role in effectively treating and preventing tuberculosis (TB), a prevalent infectious disease worldwide. 
However, TB diagnosis remains a significant challenge, particularly in resource-constrained communities and developing countries. 
The conventional gold standard test for TB necessitates a BSL-3 laboratory and is a time-consuming process, taking several months to provide definitive results, making it impractical in many settings.
Deep learning has shown promising advancements in various domains, prompting researchers to explore its potential in computer-aided TB diagnosis (CTD).
Nonetheless, the lack of annotated data has hindered the progress of deep learning in this field. 
To address this limitation, we introduce TBX11K, a large-scale TB dataset with bounding box annotations. 
This dataset not only facilitates the training of deep neural networks for CTD but also serves as the first dataset specifically designed for TB detection.

In addition to the dataset, we propose a simple yet effective framework called SymFormer for simultaneous CXR image classification and TB infection area detection. 
Leveraging the \textit{bilateral symmetry property} inherent in CXR images, SymFormer incorporates Symmetric Search Attention (SymAttention) to extract distinctive feature representations. 
Recognizing that CXR images may not exhibit strict symmetry, we introduce Symmetric Positional Encoding (SPE) to enhance the performance of SymAttention through feature recalibration.
Furthermore, to provide a benchmark for CTD research, we introduce evaluation metrics, assess baseline models adapted from existing object detectors, and launch an online challenge. 
Our experiments demonstrate that SymFormer achieves state-of-the-art performance on the TBX11K dataset, positioning it as a strong baseline for future research endeavors.
The introduction of the TBX11K dataset, the SymFormer method, and the CTD benchmark in this study are expected to significantly advance research in the field of CTD, ultimately contributing to improved detection and management of TB worldwide.

\section*{Acknowledgments}
This work was partially supported by National Key Research and Development Program of China No. 2021YFB3100800, and the National Natural Science Foundation of China under Grant 62376283.

\bibliographystyle{IEEEtran}
\bibliography{references}

\newcommand{\AddPhoto}[1]{\includegraphics[width=1in,keepaspectratio]{Authors/#1}}

\begin{IEEEbiography}[\AddPhoto{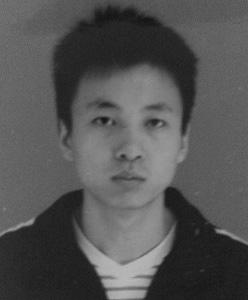}]{Yun Liu}
received his B.E. and Ph.D. degrees from Nankai University in 2016 and 2020, respectively.
Then, he worked with Prof. Luc Van Gool as a postdoctoral scholar at Computer Vision Lab, ETH Zurich, Switzerland.
Currently, he is a senior scientist at the Institute for Infocomm Research (I2R), A*STAR, Singapore.
His research interests include computer vision and machine learning.
\end{IEEEbiography}

\begin{IEEEbiography}[\AddPhoto{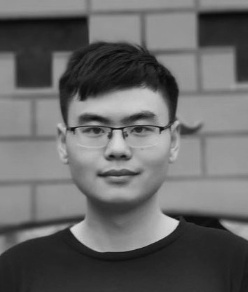}]{Yu-Huan Wu}
received his Ph.D. degree from Nankai University in 2022, advised by Prof. Ming-Ming Cheng. 
He is a scientist at the Institute of High Performance Computing (IHPC), A*STAR, Singapore.
He has published 10+ papers on top-tier conferences and journals such as IEEE TPAMI/TIP/CVPR/ICCV.
His research interests include computer vision and medical imaging.
\end{IEEEbiography}

\begin{IEEEbiography}[\AddPhoto{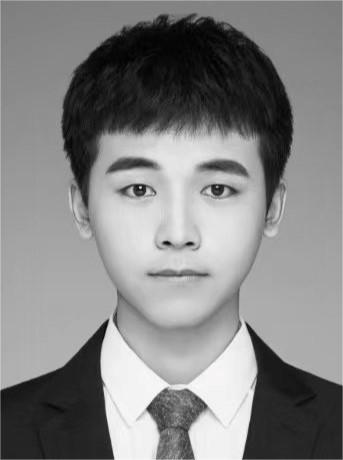}]{Shi-Chen Zhang}
received his B.E. degree in computer science
from Nankai University in 2023.
Currently, he is a Ph.D. student in Media Computing Lab, Nankai University, supervised by Prof. Ming-Ming Cheng.
His research interests include object detection and semantic segmentation.
\end{IEEEbiography}

\begin{IEEEbiography}[\AddPhoto{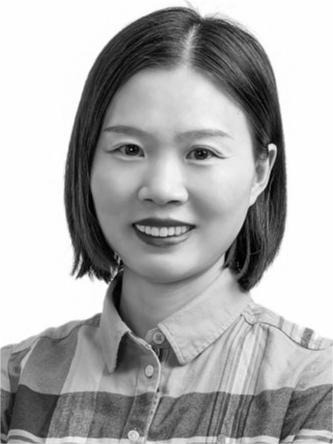}]{Li Liu}
(Senior Member, IEEE) received her Ph.D. degree from the 
National University of Defense Technology (NUDT), China, in 2012. 
She is now a Full Professor with NUDT. 
Dr. Liu served as a co-chair of many International Workshops 
along with major venues like CVPR and ICCV. 
She served as the leading guest editor of the special issues for 
IEEE TPAMI and IJCV. 
Her research interests include computer vision, pattern recognition, 
and machine learning. Her papers currently have over 10,000 citations.
\end{IEEEbiography}

\begin{IEEEbiography}[\AddPhoto{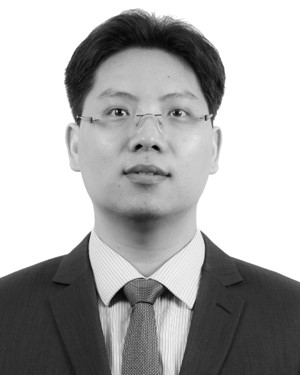}]{Min Wu}
(Senior Member, IEEE) received the B.E. degree in computer science 
from USTC, China, in 2006, 
and the Ph.D. degree in computer science from NTU, Singapore, in 2011.
He is currently a principal scientist with the Institute for Infocomm Research (I2R), A*STAR, Singapore.
He received the best paper awards in the IEEE ICIEA 2022, 
the IEEE SmartCity 2022, the InCoB 2016, and the DASFAA 2015. 
He also won the CVPR UG2+ challenge in 2021 and the IJCAI competition 
on repeated buyers prediction in 2015.
His current research interests include machine learning, data mining, 
and bioinformatics.
\end{IEEEbiography}

\begin{IEEEbiography}[\AddPhoto{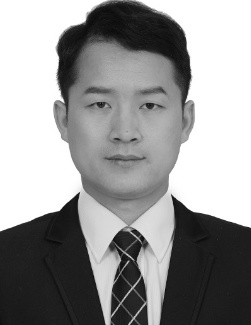}]{Ming-Ming Cheng}
received his Ph.D. degree from Tsinghua University in 2012.
Then, he did two years research fellow with Prof. Philip Torr
in Oxford.
He is now a professor at Nankai University, leading the
Media Computing Lab.
His research interests include computer graphics, computer
vision, and image processing.
He received research awards, including ACM China Rising Star Award,
IBM Global SUR Award, and CCF-Intel Young Faculty Researcher Program.
He is on the editorial boards of IEEE TPAMI/TIP.
\end{IEEEbiography}

\vfill

\end{document}